\documentclass{ieeeaccess}
\usepackage{cite}
\usepackage{amsmath,amssymb,amsfonts}
\usepackage{algorithmic}
\usepackage{graphicx}
\usepackage{textcomp}
\usepackage{amssymb}% http://ctan.org/pkg/amssymb
\usepackage{pifont}% http://ctan.org/pkg/pifont
\usepackage{hhline}
\usepackage{float}
\usepackage{hyperref}
\usepackage{array,ragged2e}
\usepackage{longtable}
 
\usepackage{multirow}

\usepackage{pgfplots,caption}
\usepackage{tikz}
  \NewSpotColorSpace{PANTONE}
  \AddSpotColor{PANTONE} {PANTONE3015C} {PANTONE\SpotSpace 3015\SpotSpace C} {1 0.3 0 0.2}
  \SetPageColorSpace{PANTONE}
\usepackage{tabularx,textcomp}
\usepackage{ctable}
\newcolumntype{C}{>{\centering\arraybackslash}X}

\def\BibTeX{{\rm B\kern-.05em{\sc i\kern-.025em b}\kern-.08em
    T\kern-.1667em\lower.7ex\hbox{E}\kern-.125emX}}
\begin{document}
% \history{Date of publication xxxx 00, 0000, date of current version xxxx 00, 0000.}
% \doi{10.1109/ACCESS.2017.DOI}

\title{Current Status and Performance Analysis of Table Recognition in Document Images with Deep Neural Networks}
\author{\uppercase{Khurram Azeem Hashmi}\authorrefmark{1,2,3},     %\IEEEmembership{Fellow, IEEE}%,
\uppercase{Marcus Liwicki\authorrefmark{4}, Didier Stricker\authorrefmark{1,2}, Muhammad Adnan Afzal\authorrefmark{5}, Muhammad Ahtsham Afzal\authorrefmark{5} and Muhammad Zeshan Afzal\authorrefmark{1,2,3}
%\IEEEmembership{Member, IEEE}%
}}

\address[1]{German Research Center for Artificial Intelligence, 67663 Kaiserslautern, Germany}
\address[2]{Department of Computer Science, University of Kaiserslautern, 67663 Kaiserslautern, Germany}
\address[3]{Mindgrage, University of Kaiserslautern, 67663 Kaiserslautern, Germany}
\address[4]{Luleå University of Technology, A3570 Luleå, Sweden}
\address[5]{Bilojix Soft Technologies, Bahawalpur. Pakistan} 
% \address[3]{Electrical Engineering Department, University of Colorado, Boulder, CO 
% 80309 USA}
% \tfootnote{This paragraph of the first footnote will contain support 
% information, including sponsor and financial support acknowledgment. For 
% example, ``This work was supported in part by the U.S. Department of 
% Commerce under Grant BS123456.''}

\markboth
{K.A. Hashmi \headeretal: Current Status and Performance Analysis of Table Recognition in Document Images with Deep Neural Networks}
{K.A. Hashmi \headeretal: Current Status and Performance Analysis of Table Recognition in Document Images with Deep Neural Networks}

\corresp{Corresponding author: Khurram Azeem Hashmi (e-mail: khurram\_azeem.hashmi@dfki.de).}

\begin{abstract}
The first phase of table recognition is to detect the tabular area in a document. Subsequently, the tabular structures are recognized in the second phase in order to extract information from the respective cells. Table detection and structural recognition are pivotal problems in the domain of table understanding. However, table analysis is a perplexing task due to the colossal amount of diversity and asymmetry in tables. Therefore, it is an active area of research in document image analysis. Recent advances in the computing capabilities of graphical processing units have enabled the deep neural networks to outperform traditional state-of-the-art machine learning methods. Table understanding has substantially benefited from the recent breakthroughs in deep neural networks. However, there has not been a consolidated description of the deep learning methods for table detection and table structure recognition. This review paper provides a thorough analysis of the modern methodologies that utilize deep neural networks. This work provided a thorough understanding of the current state-of-the-art and related challenges of table understanding in document images. Furthermore, the leading datasets and their intricacies have been elaborated along with the quantitative results. Moreover, a brief overview is given regarding the promising directions that can serve as a guide to further improve table analysis in document images.
\end{abstract}


\titlepgskip=-15pt

\maketitle

%organization image here%
\begin{figure*}[ht]
    \includegraphics[width = \linewidth]{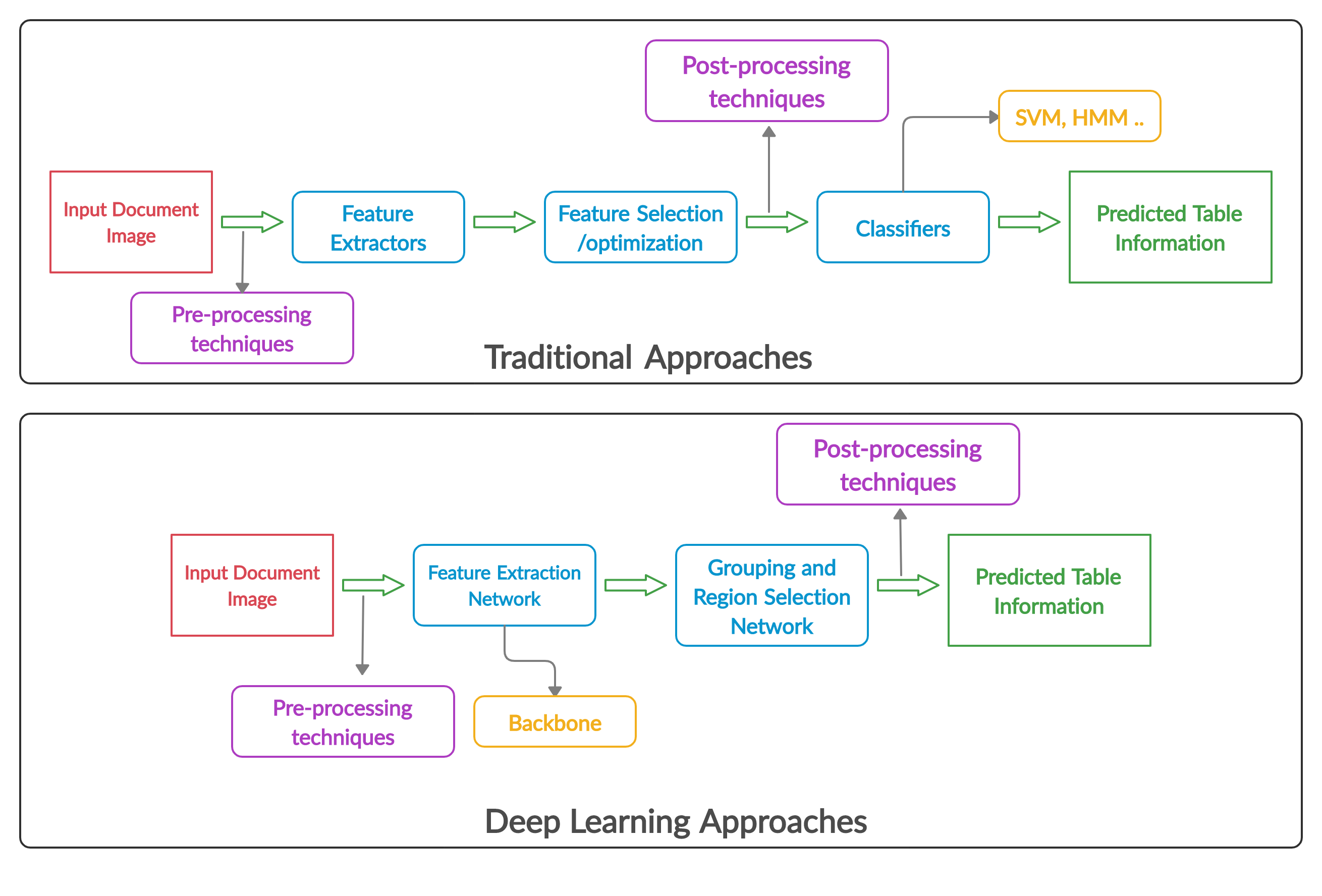}
    \caption{Pipeline comparison of traditional and deep learning approaches for table analysis. Feature extraction in traditional approaches is mainly achieved through image processing techniques whereas convolutional networks are employed in deep learning techniques. Unlike traditional approaches, deep learning methods for table understanding are not data dependent and they have better generalization capabilities. }
    \label{fig:traditional_deep}
\end{figure*}

\section{Introduction}
\label{sec:introduction}
\PARstart{T}{able} understanding has gained an immense attraction since the last decade. Tables are the prevalent means of representing and communicating structured data \cite{sarawagi2008information}. With the rise of Deep Neural Networks (DNN), various datasets for table detection, segmentation, and recognition have been published \cite{shahab2010open, fang2012dataset}. This allows the researchers to employ the DNN to improve state-of-the-art results.

Previously, the problem of table recognition has been treated with traditional approaches \cite{kim2008extracting,chen2000mining,masuda2004recognition,tyan1999generator}. One of the earlier works in the area of table analysis has been done by Kieninger \textit{et al.} \cite{kieninger1998paper,kieninger1998table,kieninger2001applying}. Along with detecting the tabular area, their system known as T-Recs extracts the structural information of the tables.

Later, machine learning techniques are applied to detect the table. One of the pioneers are Cesarini \textit{et al.} \cite{cesarini2002trainable}. Their proposed system, Tabfinder converts a document into an MXY tree which is a hierarchical representation of the document. It searches for a block region in the horizontal and vertical parallel lines, and then a depth-first search to handle noisy document images leads to a tabular region. Silva \textit{et al.} \cite{e2009learning} adopted rich Hidden-Markov-Models to detect tabular area based on joint probability distributions.

%organization image here%
\begin{figure*}[ht]
    \includegraphics[width = \linewidth]{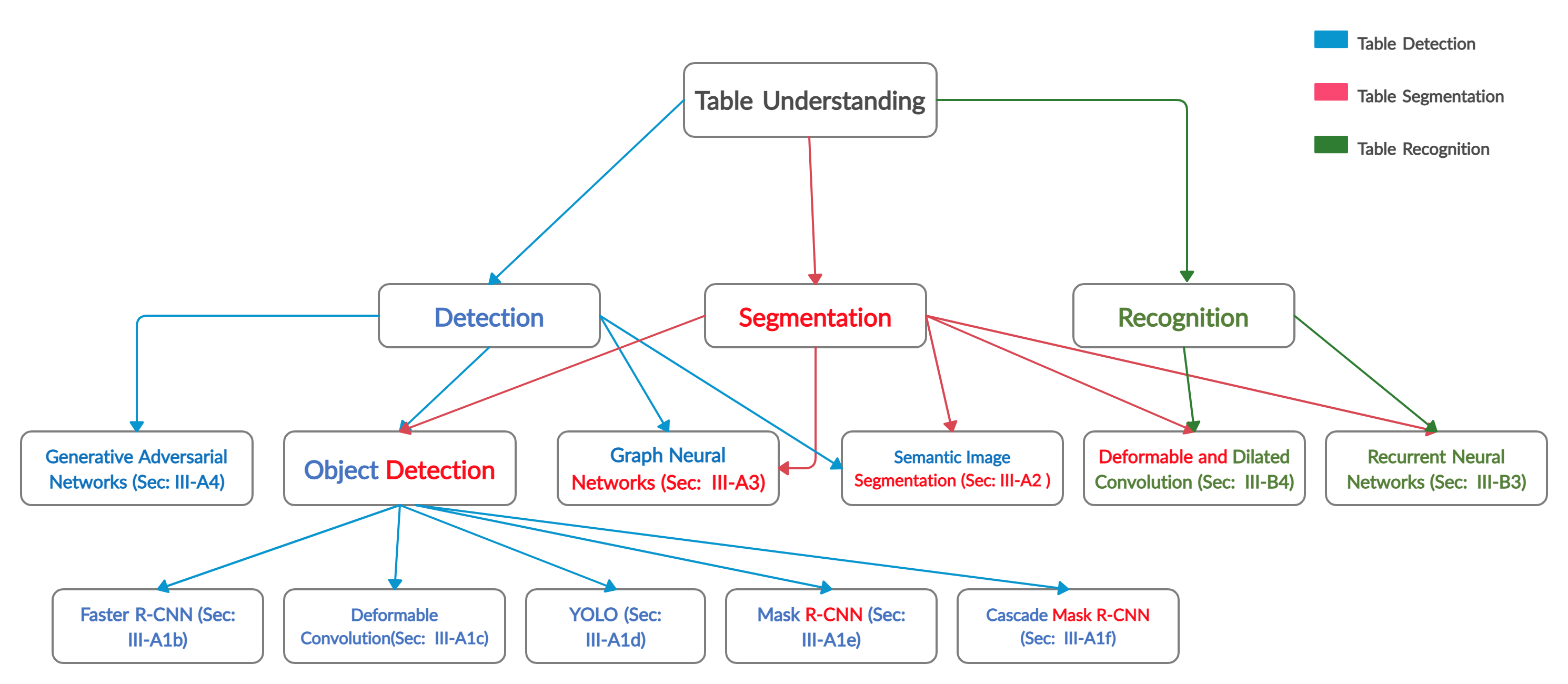}
    \caption{Organization of explained methodologies in the paper. Concepts written in blue color represent table detection techniques. Methods in red color demonstrate the table segmentation or table structure recognition approaches, whereas the architectures in green color depict the table recognition method, which involves the extraction of cell content in a table. As illustrated, some of the architectures have been exploited in multiple tasks of table understanding.}
    \label{paper_organization}
\end{figure*}

Support Vector Machines (SVM) \cite{cortes1995support} have also been exploited along with some handcrafted features to detect tables \cite{kasar2013learning}. \textit{Fan et al.} \cite{fan2015detecting} tried to detect tables by the fusion of various classifiers trained on linguistic and layout information of documents. Another work carried out by \textit{Tran et al.} \cite{tran2015table} uses a region of interest to detect tables in document images. These regions are further filtered as tables if the text block present in the region of interest satisfies a specific set of rules.

Comprehensive research is conducted by \textit{Wang et al.} \cite{wang2004table} focusing not only on the problem of table detection but table decomposition as well. Their probability optimization-based algorithm is similar to the well-known X-Y cut algorithm \cite{nagy1984hierarchical}. The system published by \textit{Shigarov et al.} \cite{shigarov2016configurable} leverages the bounding boxes of words to restore the structure of a table. Since the system is heavily dependent on the metadata, the authors have employed PDF files to execute this experiment.

Figure \ref{fig:traditional_deep} depicts the standard pipeline comparison between traditional approaches and deep learning methods for the process of table understanding. Traditional table recognition systems are either not generic enough on different datasets or they require the additional metadata from PDF files. In most of the traditional methods, exhaustive pre and post-processings were also employed to enhance the performance of traditional table recognition systems. However, in deep learning systems, instead of handcrafted features, neural networks mainly convolutional neural networks \cite{krizhevsky2012imagenet} are used to extract features. Subsequently, object detection or segmentation networks attempt to distinguish the tabular part which is further decomposed and recognized in a document image. 

The text documents can be classified into two categories. The first category belongs to born-digital documents that contain not only text but the related meta-data such as layout information. One such example is the PDF documents. The second category of documents is acquired using devices such as scanners and cameras. To the best of our knowledge, there is no notable work that has employed deep learning for table recognition in camera-captured images. However, in the literature, one heuristic based approach \cite{seo2015junction} exists that works with camera-captured document images. The scope of this survey is to assess the deep learning-based approaches that have performed table recognition on the scanned document images. 

This review paper is organized as follows: Section \ref{sec:related_work} provides a brief discussion about the reviews and surveys which are already published in the research community; Section \ref{sec:methodology} provides an exhaustive discussion about several approaches that have contributed in the area of table understanding by leveraging deep learning concepts. Figure \ref{paper_organization} explains the structural flow of mentioned methodologies; Section \ref{sec:datasets} investigates all the publicly available datasets that can be exploited for the problems in table analysis; Section \ref{sec:evaluation} explains the well known evaluation metrics and provides performance analysis of all the discussed approaches in Section \ref{sec:methodology}; Section \ref{sec:conclusion} concludes the discussion where as Section\ref{sec:futework} highlights various open issues and future directions.

\section{RELATED WORK}
\label{sec:related_work}
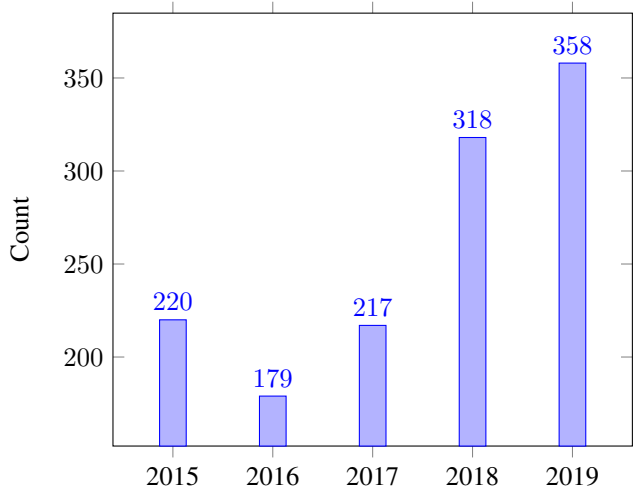
\begin{figure}
    \begin{tikzpicture}
        \begin{axis}[
            title=\textbf{ Annual Number of Publications in Table Analysis },
            ybar,
            enlargelimits=0.15,
            legend style={at={(0.5,-0.15)},
              anchor=north,legend columns=-1},
            ylabel={Count},
            symbolic x coords={2015, 2016,2017,2018,2019},
            xtick=data,
            nodes near coords,
            nodes near coords align={vertical},
            ]
        \addplot coordinates {(2015,220) (2016,179) (2017,217) (2018,318) (2019,358)};
        \end{axis}
    \end{tikzpicture}
\captionof{figure}{ Illustration of an increasing trend in the domain of table analysis. This data is collected by checking the yearly publications on table detection and table recognition from year 2015 to the year 2019. }
\label{graph:pub_count}
\end{figure}

The problem of table analysis has been a well-recognized problem for several years. Figure \ref{graph:pub_count} illustrates the increasing trend in the number of publications for the last 5 years. Since this is a review paper, we would like to shed some light on the previous surveys and reviews that are already available in the table community. In the chapter \textit{Document Recognition} in one of his books, Dougherty defines table \cite{dougherty1999electronic}. In the survey on document recognition, Handley \cite{handley2000table} elaborated on the task of table recognition along with a precise explanation of previous work done in this domain. Later, \textit{Lopresti et al.} \cite{lopresti1999automated} presented the survey on table understanding in which they discussed the heterogeneity in different kinds of tables. They also pointed out the potential areas where improvement could be made by leveraging many examples. The comprehensive survey was transformed into a tabular form which was later published as a book \cite{lopresti1999tabular}. 

\textit{Zanibbi et al.} \cite{zanibbi2004survey} came up with the exhaustive survey which includes all the recent material and state-of-the-art approaches of that time. They define the problem of table recognition as "the interaction of models, observations, transformations, and inferences"\cite{embley2006table}. Hurst in his doctoral thesis \cite{hurst2000interpretation} defines the interpretation of tables. \textit{Silva et al.} \cite{e2006design} published another survey in 2006. Along with evaluating the current table processing algorithms, the authors have proposed their own end-to-end table processing method and evaluation metrics to solve the problem of table structure recognition. 

\textit{Embley et al.} \cite{embley2006table} wrote a review illustrating about the table-processing paradigms.In 2014, another review on table recognition and forms is published by \textit{Coüasnon et al.}\cite{couasnon2014recognition}. The review covers a brief overview of the recent approaches of that time. In the following year and according to our knowledge, the latest review on the detection and extraction of tables in PDF documents is published by \textit{Khusro et al.}\cite{khusro2015methods}. 

\section{METHODOLOGIES}
\label{sec:methodology}
As elaborated in \cite{zhong2019image}, we have also defined the problem of table understanding into three steps:

\begin{enumerate}
    \item \textit{Table Detection:} detecting the tabular boundaries in terms of bounding boxes in document images.
    \item \textit{Table Structural Segmentation:} defines the structure of table by analyzing information of row and column layouts.
    \item \textit{Table Recognition:} includes both structural segmentation and parsing information of table cells.
\end{enumerate}

% 1) \textit{table detection:} detecting the tabular boundaries in terms of bounding boxes in document images, 2) \textit{table structural segmentation:} defines the structure of table by analyzing information of row and column layouts, 3) \textit{table recognition:} includes both structural segmentation and parsing information of table cells.

\begin{figure*}[ht]
    \includegraphics[width = \linewidth]{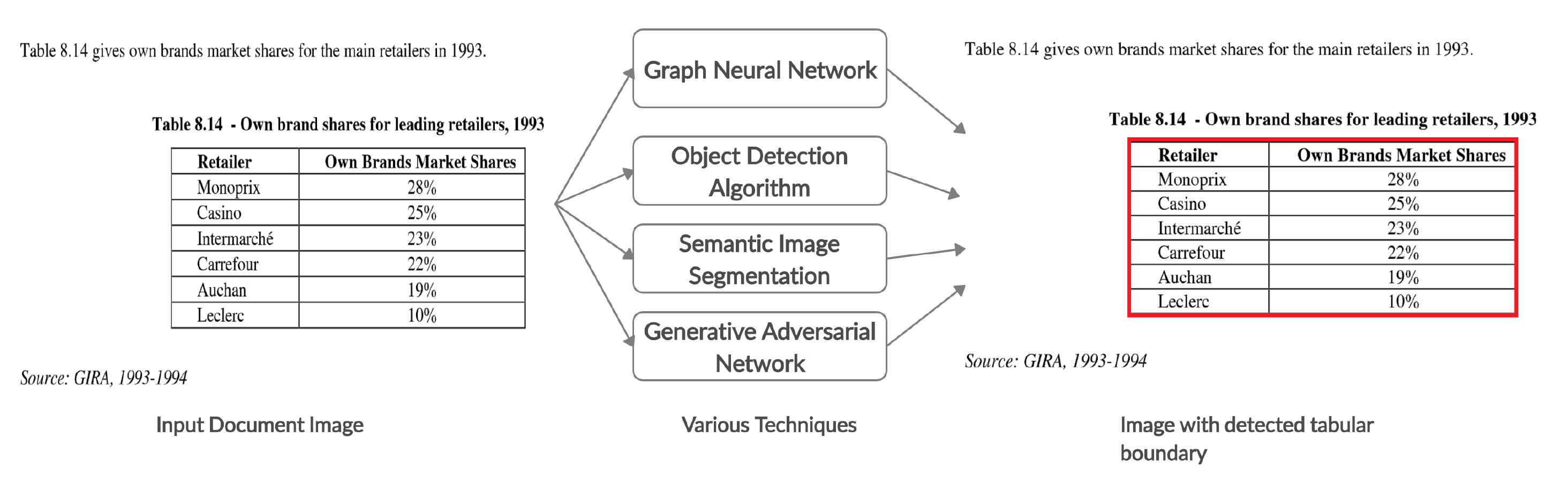}
    \caption{Basic flow of table detection along with the methods used in the discussed approaches. In order to locate the tabular boundaries, document image is passed through various deep learning architectures.}
    \label{fig_detection}
\end{figure*}

\subsection{Table Detection}
\label{sec:table_detect}
The first part of extracting information from the tables is to identify the tabular boundary in the document images \cite{hu2002evaluating}. Figure \ref{fig_detection} explains the fundamental flow of table detection which has been discussed in numerous approaches. Various deep learning concepts have been employed to detect tabular areas from the document images. This section reviews the deep learning techniques which are exploited to perform table detection in document images. For the sake of providing convenience to our readers, we have categorized the approaches into discrete deep learning concepts. Table \ref{tab:comparison-faster-rcnn} summarizes all the object detection-based table detection approaches,whereas Table \ref{tab:comparison-remaining-table-det} highlights the advantages and limitations of the methods that have applied other deep learning-based techniques.

Based on our knowledge, the first approach that employed deep learning methods to solve the table detection task is proposed by \textit{Hao et al.}\cite{hao2016table}. Along with the use of a convolutional neural network to extract the image features, authors applied some heuristics by leveraging the PDF metadata. Since this technique is based on PDF documents rather than relying on document images, we decide not to include this research in our performance analysis.

\subsubsection{Object Detection Algorithms}

Object detection is a branch of deep learning which deals with detecting an object in any image or a video frame. Region-based object detection algorithms are mainly divided into two steps: the first one is to generate appropriate proposals also known as \textit{region of interest}. These regions of interest are classified using convolutional neural networks in the second step.

\paragraph{Transfer Learning}
Transfer learning is the concept of utilizing a pre-trained model on a problem that belongs to a different, but related domain \cite{torrey2010transfer}. Due to limited number of available labelled datasets, transfer learning has been excessively used in the vision-based approaches \cite{tl_1,tl_2,tl_3,tl_4}. For similar reasons, researchers in the document image analysis community have also powered the capabilities of transfer learning to advance their approaches \cite{loey2020within,afzal2015deepdocclassifier,das2018document}. The capabilities of transfer learning have aided the researchers to reuse the pre-trained networks (trained on ImageNet \cite{krizhevsky2012imagenet} or COCO \cite{lin2014microsoft}) on the problem of table detection and table structure recognition in document images \cite{gilani2017table,schreiber2017deepdesrt,siddiqui2018decnt,huang2019yolo,prasad2020cascadetabnet,agarwal2020cdec, siddiqui2019deeptabstr,hashmi2021guided, zheng2021global, raja2020table}. While Section \ref{sec:fasterrcnn}, \ref{sec:deformconv} and \ref{sec:cascade} explains transfer learning-based table detection methods, the techniques that employed transfer learning for the task of table structure recognition are elaborated in Section \ref{sec:odstr}. 

\paragraph{Faster R-CNN}
\label{sec:fasterrcnn}

After the improvement of object detection algorithms from Fast R-CNN \cite{girshick2015fast} to Faster R-CNN \cite{ren2015faster}, the tables are treated as an object in the document images. Gilani \textit{et al.} \cite{gilani2017table} employed deep learning method on the images to detect tables. The technique involves image transformation as a pre-processing step that follows with the table detection. In the image transformation part, a binary image is taken as an input on which Euclidean distance transform \cite{breu1995linear}, linear distance transform \cite{fabbri20082d}, and max distance transform \cite{ragnemalm1993euclidean} are applied on blue, green and red channels of the image respectively. Later, Gilani \textit{et al.} \cite{gilani2017table} have used a region-based object detection model called Faster R-CNN \cite{ren2015faster}. The backbone of their Region Proposal Network (RPN) is based on ZFNet \cite{zeiler2014visualizing}. Their approach was able to beat the state-of-the-art results on UNLV \cite{shahab2010open} dataset. 

One of the works executed on document images by using the capabilities of deep learning has been accomplished by Schreiber \textit{et al.} \cite{schreiber2017deepdesrt}. Their end to end system known as \textit{DeepDeSRT} not only detects the tabular region but also distinguishes the structure of the table and both of these tasks are dealt with by applying distinctive deep learning techniques. Table detection has been achieved by using Faster R-CNN \cite{ren2015faster}. They have experimented with two different architectures as their backbone network:  Zeiler and Fergus (ZFNet) \cite{zeiler2014visualizing} and a deep VGG-16 network \cite{simonyan2014very}. Models are pre-trained on Pascal VOC \cite{everingham2010pascal} dataset. Method for structural segmentation is explained in the Section \ref{sec:segmentation}
% one of three famous object detection datasets: ImageNet \cite{b18}, Microsoft COCO \cite{b19} or Pascal VOC \cite{b20}. %

With an increase in memory of graphical processing units (GPU), a room for bigger public datasets is created to completely leverage the power of GPUs. Minghao \textit{et al.} \cite{li2020tablebank} comprehends this need and proposed \textit{TableBank}, which contains 417K labeled tables and their respective document images. They have also suggested baseline models by using Faster R-CNN \cite{ren2015faster} for the task of table detection. The author proposed a baseline method for structure recognition as well which will be explained later in Section \ref{sec:segmentation}

In another research presented in the ICDAR 2019 conference, tables are detected using the combination of Faster R-CNN and further improved using the locating corners method  \cite{sun2019faster}. The authors define the corners like a square of size $80 \times 80$ drawn around the vertices of tables. Along with locating the boundary of tables, corners are also detected using the same Faster R-CNN model. These corners are further refined after passing through various heuristics like two consecutive corners are on the same horizontal line. After analyzing the corners, inaccurate corners are filtered and left to form a group. The authors argue that most of the time, inaccuracy in table boundaries is due to inaccurate detection of the left and right side of the boundaries as compared to the top and bottom side of boundaries. Hence, only the right and left sides of a detected table are refined in this experiment. The refinement is carried out by first finding the corresponding corner for a table by calculating the intersection over the union between them. Subsequently, horizontal points of the table are shifted by taking the mean value between the table boundary and the corresponding corner. This article has conducted an experiment on ICDAR 2017 page object detection dataset \cite{icdar17} and reported a 2.8\% increase in F-measure as compared to the traditional Faster R-CNN approach.

\begin{paragraph}{Deformable Convolutions}
\label{sec:deformconv}
Another approach is proposed by Siddiquie \textit{et al.} \cite{siddiqui2018decnt} in 2018 which was a follow-up work of Schreiber \textit{et al.} \cite{schreiber2017deepdesrt}. They have performed the table detection tasks by taking advantage of deformable convolutional neural networks \cite{dai2017deformable} in the model of Faster R-CNN. The authors claim that deformable convolutions exceeds the performance of traditional convolutions due to having various tabular layouts and scales in the documents. Their model \textit{DeCNT} have shown state-of-the-art results on the datasets of ICDAR-2013 \cite{icdar13}, ICDAR-2017 POD \cite{icdar17}, UNLV \cite{shahab2010open} and Marmot \cite{fang2012dataset}.
% \Figure[t!](topskip=0pt, botskip=0pt, midskip=0pt){decnt.png}
% {Magnetization as a function of applied field.
% It is good practice to explain the significance of the figure in the caption.\label{decnt}}
\end{paragraph}

Agarwal \textit{et al.} \cite{agarwal2020cdec} presented the approach called CDeC-Net (Composite Deformable Cascade Network) to detect tabular boundaries in document images. In this work, the authors empirically established that there is no need to add extra pre/post-processing techniques to obtain state-of-the-art results for table detection. This work is based on a novel cascade Mask R-CNN \cite{cai2018cascade} along with the composite backbone which is a dual backbone architecture (two ResNeXt-101 \cite{xie2017aggregated}) \cite{liu2020cbnet}. In their composite backbone, the authors replace the conventional convolutions with the deformable convolutions to address the problem of detecting tables with arbitrary layouts. With the combination of deformable composite backbone and strong Cascade Mask R-CNN, their proposed system produced comparable results on several publicly available datasets in the table community.

\paragraph{YOLO}
\label{sec:YOLO}
YOLO (You Only Look Once) \cite{redmon2016you} which is a famous model for detecting objects in real-world images efficiently has also been employed in the task of table detection by Huang \textit{et al.} \cite{huang2019yolo}. YOLO is different from region proposal methods because it handles the task of object detection more like a regression instead of a classification problem. YOLOv3 \cite{redmon2018yolov3} is the recent and enhanced version of YOLO \cite{redmon2016you} and is therefore used in this experiment. In order to make the predictions more precise, white-space margins are removed from the predicted tabular area along with the refinement of noisy page objects.

\paragraph{Mask R-CNN, YOLO, SSD and Retina Net}
\label{sec:mask-yolo-ssd}
Another research that leverages object detection algorithms is "The Benefits of Close-Domain Fine-Tuning for Table Detection in Document Images" published by \textit{Casado-García et al.} \cite{casado2020benefits}. After carrying out an exhaustive evaluation, the authors have demonstrated the improvement in the performance of table detection when fine-tuned from a closer domain.  Leveraging the object detection algorithms, the writers have used Mask R-CNN \cite{he2017mask}, YOLO \cite{deng2016you}, SSD \cite{liu2016ssd} and Retina Net \cite{lin2017focal}. To conduct this experiment, two base datasets are selected. The first dataset was PascalVOC \cite{everingham2010pascal} which contains natural scenic images and has no close relation with the datasets present in the table community. The second base dataset was TableBank \cite{li2020tablebank} which has 417 thousand labeled images further explained in Section \ref{sec:TableBank}. Two separate models were trained on these datasets and tested comprehensively on all ICDAR table competitions datasets along with other datasets like Marmot and UNLV \cite{shahab2010open} which are later explained in Section \ref{sec:datasets}. An average of 17\% in improvement is noted in this article when models are fine-tuned with closer domain datasets as compared to models trained on real-world images.

\paragraph{Cascade Mask R-CNN}
\label{sec:cascade}
Along with the recent improvements in generic spatial feature extraction networks \cite{wang2020deep, gao2019res2net}, and object detection networks \cite{cai2018cascade,chen2019hybrid}, we have seen a noticeable improvement in table detection systems. Prasad \textit{et al.} \cite{prasad2020cascadetabnet} published the \textit{CascadeTabNet} which is an end-to-end table detection and structure recognition method. In this work, the authors leverage the novel blend of Cascade Mask R-CNN \cite{cai2018cascade} (which is a multistage Mask R-CNN) with the HRNet \cite{wang2020deep} as a base network. The paper exploited the similar area proposed by \cite{gilani2017table} and instead of raw document images, transformed images were fed to the strong Cascade Mask R-CNN \cite{cai2018cascade}. Their proposed system was able to achieve state-of-the-art results on the datasets of ICDAR-2013 \cite{icdar13}, ICDAR-2019 \cite{icdar19} and TableBank \cite{li2020tablebank}.

In one of the very recent works, Zheng \textit{et al.} \cite{zheng2021global} published a framework for both the detection and structure recognition of tables in document images. The authors argue that the proposed system \textit{GTE} (Global Table Extractor) is a generic vision-based method in which any object detection algorithm can be employed. The method feeds raw document images to the multiple object detectors that simultaneously detect tables and the individual cells to achieve accurate table detection. The predicted tables by the object detectors are further refined with the help of an additional penalty loss and predicted cellular boundaries. The approach further improves the predicted cellular areas to tackle table structure recognition, and it is explained in Section \ref{sec:segmentation}.

\subsubsection{Semantic Image Segmentation}
\label{sec:sis}
In the year 2018, the combination of deep convolutional neural networks, graphical models, and the concepts of saliency features have been applied to detect charts and tables by Kavasidis \textit{et al.} \cite{kavasidis2018saliency}. The authors argued that instead of using the object detection networks, the task of detecting the tables can be posed as a saliency detection. The model is based on a semantic image segmentation technique. It first extracts saliency features and then each pixel is classified whether that pixel belongs to a region of interest or not. To notice long-term dependencies, the model employed dilated convolutions \cite{yu2016multi}. In the end, the generated saliency map is propagated to the fully connected Conditional Random Field (CRF) \cite{krahenbuhl2011efficient}, which further improves the predictions.

\paragraph{Fully Convolutional Networks}
\label{sec:fcn}
\textit{TableNet} powered by deep learning is an end-to-end model for both detecting as well as recognizing the structure of tables in document images presented by Shubham et al \cite{paliwal2019tablenet}. The proposed method exploits the concepts of fully convolutional networks \cite{long2015fully} with a pre-trained VGG-19 \cite{simonyan2014very} layer as the base network. The author claims that the problem of identifying the tabular area and structure recognition can be jointly addressed similarly. They further demonstrated how the performance of a new dataset can be enhanced by exploiting the capabilities of transfer learning.

\begin{figure}
    \begin{tikzpicture}
        \begin{axis}[
            title=\textbf{Deep Learning Methods used in Table Detection },
            ybar,
            enlargelimits=0.15,
            legend style={at={(0.5,-0.15)},
              anchor=north,legend columns=-1},
            ylabel={Count},
            symbolic x coords={OD,GNN,GAN,SIS},
            xtick=data,
            nodes near coords,
            nodes near coords align={vertical},
            ]
        \addplot coordinates {(OD,10) (GNN,2) (GAN,1) (SIS,2)};
        \end{axis}
    \end{tikzpicture}
\captionof{figure}{OD is Object Detection, SIS means Semantic Image Segmentation, GNN is Graph Neural Networks whereas GAN is used to represent Generative Adversarial Networks. This graph explains what kind of deep learning algorithms are periodically exploited to perform table detection.}
\label{graph:detect}
\end{figure}
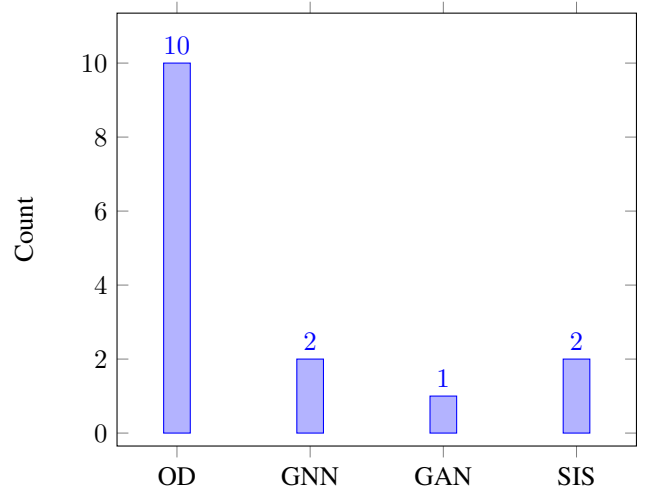

\begin{table*}
\centering

\caption{ A summary of advantages and limitations of various deep learning-based table detection methods that are based on object detection frameworks.}
\normalsize
\setlength\extrarowheight{5pt}
\newcolumntype{l}{>{\hsize=.3\hsize}X}
\newcolumntype{m}{>{\hsize=.5\hsize}X}
\newcolumntype{a}{>{\hsize=.5\hsize}X}
\newcolumntype{b}{>{\hsize=.5\hsize}X}

\begin{tabularx}{\textwidth}{>{\justifying\arraybackslash\noindent}l|m|a|b}
\specialrule{.2em}{.1em}{.1em} 
\captionsetup{font=scriptsize}
\textbf{Literature}&
\textbf{Method}&
\textbf{Highlights}&
\textbf{Limitations}\\
% \hline
    \specialrule{.2em}{.1em}{.1em}
    \footnotesize \textit{ Gelani et al.} \cite{gilani2017table}& 
    \footnotesize Faster R-CNN (Section \ref{sec:fasterrcnn}).
    
    Images are transformed and then fed into the Faster R-CNN. & 
    \footnotesize 
    \textbf{a)} First deep learning based table detection approach on scanned document images, \textbf{b)} Transforming RGB pixels to distance metrics facilitates the object detection algorithm. & \footnotesize 
    Extra pre-processing steps involved. \footnotesize \\
    \hline
    \footnotesize \textit{DeCNT} \cite{siddiqui2018decnt}& \footnotesize 
    Deformable convolutions implemented in the Faster R-CNN architecture (Section \ref{sec:deformconv}). & 
    \footnotesize The dynamic receptive field of deformable convolutional neural networks help in recognizing various tabular boundaries. & 
    \footnotesize Deformable convolutions are computationally intensive as compared to traditional convolutions.\\
    \hline
    \footnotesize \textit{DeepDeSRT} \cite{schreiber2017deepdesrt}& 
    \footnotesize Faster R-CNN with transfer learning techniques (Section \ref{sec:fasterrcnn})& \footnotesize Simple and effective end-to-end approach to detect tables and structures of the tables.& 
    \footnotesize Not as accurate as compared to other states of the art approaches. \\
    \hline
    \footnotesize \textit{TableBank} \cite{li2020tablebank}& \footnotesize
    Faster R-CNN used as a baseline method for a novel dataset (Section \ref{sec:fasterrcnn}). & \footnotesize 
    This approach presents that by leveraging a large dataset such as TableBank, a simple Faster R-CNN can produce impressive results. & \footnotesize 
    Just a direct application of Faster R-CNN. \\
    \hline
    \footnotesize \textit{Sun et al.} \cite{sun2019faster}& \footnotesize  
    Faster R-CNN with locating corners (Section \ref{sec:fasterrcnn}). & \footnotesize 
    \textbf{a)} Faster R-CNN is exploited to detect not only tables but the corners of the tabular boundaries as well, \textbf{b)} Novel method produces better results.  & \footnotesize  
    \textbf{a)} Computationally more extensive because of additional detections, \textbf{b)} Post-processing steps such as corners' refinement are required. \\
    
    \hline
    \footnotesize \textit{Huang et al.} \cite{huang2019yolo}& \footnotesize 
    YOLO based table detection method (Section \ref{sec:YOLO}).& 
    \footnotesize Comparatively, faster and efficient approach. & 
    \footnotesize The proposed method depends on the data driven post-processing techniques. \\
    \hline
    \footnotesize \textit{García et al.} \cite{casado2020benefits} &
    \footnotesize Employed Mask R-CNN, YOLO, SSD and RetinaNet to compare fine-tuning techniques (Section \ref{sec:mask-yolo-ssd}). & \footnotesize 
    Presented the benefits of leveraging a closer domain fine-tuning methods for table detection while employing object detection networks. & 
    \footnotesize Still, closed domain fine-tuning is not enough to reach the state-of-the-art results. \\
    \hline
    \footnotesize \textit{CascadeTabNet} \cite{prasad2020cascadetabnet} &
    \footnotesize Employed Cascade Mask R-CNN with an iterative transfer learning approach (Section \ref{sec:cascade}). & \footnotesize 
    This work presents that transformed images with an iterative transfer learning can reduce the dependency of large-scale datasets. & 
    \footnotesize Similar to \cite{gilani2017table}, extra pre-processing steps are involved in this approach.\\
    \hline
    \footnotesize \textit{CDeC-Net} \cite{agarwal2020cdec} &
    \footnotesize Cascade Mask R-CNN with a deformable composite backbone (Section \ref{sec:deformconv}). & \footnotesize 
    \textbf{a)} Extensive evaluations on publicly available benchmark datasets for table detection. \textbf{b)} An end-to-end object detection-based framework leveraging composite backbone to produce state-of-the-art results.& 
    \footnotesize Along with the deformable convolutions, a composite backbone is employed which makes the approach computationally intensive.\\
    \hline
    \footnotesize \textit{GTE} \cite{zheng2021global} &
    \footnotesize Proposed a generic object detection approach (Section \ref{sec:cascade}). & \footnotesize \textbf{a}) An end-to-end technique that can operate on any object detection framework. \textbf{b}) This work proposed an additional piece-wise constraint loss that benefits the task of table detection. & 
    \footnotesize Since the task of table detection is dependent on cell detections, annotations for cellular boundaries are required.\\
    \specialrule{.2em}{.1em}{.1em}

\end{tabularx} 
\label{tab:comparison-faster-rcnn}
\end{table*}

%% NEW TABLE FOR REMAINING TABLE DETECTION APPROACHES BEGIN
\begin{table*}
\centering

\caption{ A summary of advantages and limitations of various table detection methods that operate on other deep learning-based concepts. The bold horizontal line separates the approaches with different architectures. }
\normalsize
\setlength\extrarowheight{5pt}
\newcolumntype{l}{>{\hsize=.3\hsize}X}
\newcolumntype{m}{>{\hsize=.5\hsize}X}
\newcolumntype{a}{>{\hsize=.5\hsize}X}
\newcolumntype{b}{>{\hsize=.5\hsize}X}

\begin{tabularx}{\textwidth}{>{\justifying\arraybackslash\noindent}l|m|a|b}
\specialrule{.2em}{.1em}{.1em} 
\captionsetup{font=scriptsize}
\textbf{Literature}&
\textbf{Method}&
\textbf{Highlights}&
\textbf{Limitations}\\
% \hline
    \specialrule{.2em}{.1em}{.1em}
    \footnotesize \textit{Kavasidis et al.} \cite{kavasidis2018saliency} &
    
    \footnotesize Semantic Image Segmentation with saliency concepts (Section \ref{sec:sis}). & 
    \footnotesize \textbf{a)} This method poses the task of table detection as saliency detection, \textbf{b)} Dilated convolutions are applied instead of traditional convolutions.& 
    \footnotesize Multiple processing steps are required to achieve comparable results. \\
    \hline
    \footnotesize \textit{TableNet} \cite{paliwal2019tablenet} &
    \footnotesize Fully Convolutional Networks (Section \ref{sec:fcn}). & 
    \footnotesize \textbf{a)} An end-to-end approach for table detection and structure recognition in document images, \textbf{b)} First approach to jointly address the task of table detection and structure recognition with a single method.& 
    \footnotesize In the case of table structural extraction, this technique only works on column detection. \\
    \specialrule{.2em}{.1em}{.1em}
    % \specialrule{.2em}{.1em}{.1em}
    \footnotesize \textit{Martin et al.} \cite{holevcek2019table} &
    \footnotesize Graph Neural Network with the line item detection approach. (Section \ref{sec:gnn}) & 
    \footnotesize The method shows promising results on the layout-heavy documents such as invoices. & 
    \footnotesize \textbf{a)} Approach is not evaluated on any publicly available table datasets, \textbf{b)} Weak baseline method and no comparisons with other state-of-the-art methods.\\ 
    \hline
    \footnotesize \textit{Riba et al.} \cite{riba2019table} &
    \footnotesize Graph Neural Network by leveraging textual attributes through OCR (Section \ref{sec:gnn}) & 
    \footnotesize The proposed method leverages more information than just the spatial features. & 
    \footnotesize \textbf{a)} This method requires extra annotations apart from the information of tabular area, \textbf{b)} No comparisons with other state-of-the-art approaches. \\
    \hline
    \specialrule{.2em}{.1em}{.1em}
    % \specialrule{.2em}{.1em}{.1em}
    \footnotesize \textit{Li et al.} \cite{li2019gan} &
    \footnotesize Generative Adversarial Networks and object detection network (Section \ref{sec:gan}) & 
    \footnotesize GAN based approach forces the network to extract similar features for ruling and less-ruled tables. & 
    \footnotesize Model with generators is vulnerable in document images having diverse tabular layouts. \\
    \specialrule{.2em}{.1em}{.1em}

\end{tabularx} 
\label{tab:comparison-remaining-table-det}
\end{table*}

%% NEW TABLE FOR REMAINING TABLE DETECTION APPROACHES ENDSS

\subsubsection{Graph Neural Networks}
\label{sec:gnn}
Recently, we have seen that the adoption of graph neural networks in the area of table understanding is on the rise. Riba \textit{et al.} \cite{riba2019table} carried out an experiment of detecting tables using graph neural networks in the invoice documents. Due to the limited amount of information available in the images of invoices, the authors argue that graph neural networks are a better fit to detect the tabular area. The paper also publishes the labeled subset of the original RVL-CDIP dataset \cite{harley2015evaluation} which is pulbicly available.

Martin \textit{et al.} \cite{holevcek2019table} extends the application of graph neural networks by presenting the idea of table understanding using graph convolutions in structured documents like invoices. The proposed research is also conducted on PDF documents however, the authors claim that the model is robust enough to handle other kinds of data sets. In this research, the problem of table detection is solved by combining the task of line item table detection and information extraction. With the line item approach, any word can be easily distinguished whether it is a part of a line item or not. After classifying all words, the tabular area can be efficiently detected since lines in the table separate reasonably well enough as compared to other text-areas in invoices. 

\begin{figure*}
    \includegraphics[width = \linewidth]{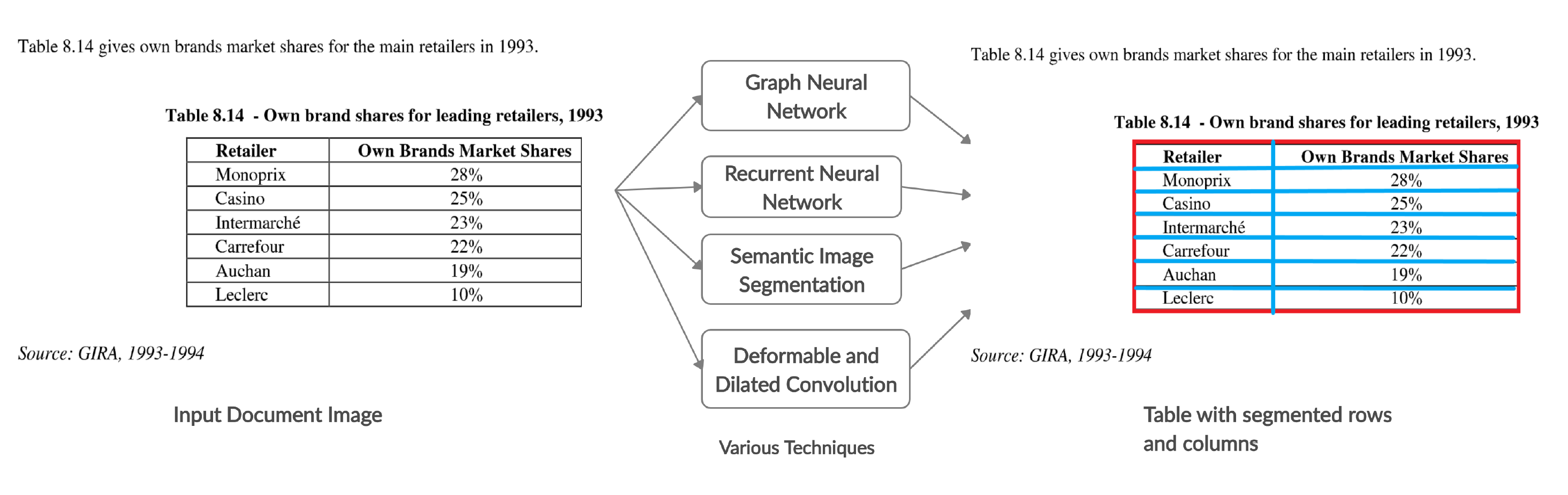}
    \caption{Basic flow of table structural segmentation along with the methods used in the discussed approaches. Instead of a document image, tabular image is given to the various deep neural architectures in order to recognize the structure of table. }
    \label{fig_segment}
\end{figure*}

\subsubsection{Generative Adversarial Networks}
\label{sec:gan}
Generative Adversarial Networks (GAN) \cite{goodfellow2014generative} have also been exploited to identify tables. The proposed approach \cite{li2019gan} makes sure that the generative network sees no difference between the ruling and less-ruling tables and try to extract identical features in both of the cases. Subsequently, the feature generator is joined with semantic segmentation models like Mask R-CNN \cite{he2017mask} or U-net \cite{ronneberger2015u}. After combining the GAN-based feature generator with Mask R-CNN, the approach is evaluated on the ICDAR2017 POD dataset \cite{icdar17}. Authors claim that this approach will facilitate other object detection and segmentation problems.

\subsection{Table Structural Segmentation}
\label{sec:segmentation}

Once, the boundary of the table is detected, the next step is to identify the rows and columns \cite{e2006design}. In this section, we will review the recent approaches that have attempted the problem of table structural segmentation. We have categorized the methodologies according to the architecture of deep neural networks. 
Table \ref{tab:comparison-segmentation} summarizes these approaches by highlighting their advantages and limitations. Figure \ref{fig_segment} illustrates the essential flow of table structural segmentation techniques that are discussed in this review paper.

\subsubsection{Semantic Image Segmentation}

Along with table detection, \textit{TableNet} segments the structure of a table by detecting the columns in respective tables. Shubham et al \cite{paliwal2019tablenet} used a pre-trained VGG-19 \cite{simonyan2014very} as a base network that acts as an encoder while a decoder performs the column detection. The author tries to convince the readers that due to the interdependence between the table detection and structural segmentation, both of the problems can be solved efficiently by using a single network.

\paragraph{Fully Convolutional Networks}
\label{sec:fcn_seg}
To recognize the structure in tables, the authors of \textit{DeepDeSRT} \cite{schreiber2017deepdesrt} have exploited the concept of semantic segmentation. They implemented a fully convolutional network proposed in \cite{long2015fully}. An added pre-processing step of stretching the table vertically for rows and horizontally for columns have provided a valuable advantage in the results. They achieved state-of-the-art results on the ICDAR 2013 table structure recognition dataset \cite{icdar13}.

Another paper "Rethinking Semantic Segmentation for Table Structure Recognition in Documents" is proposed by Siddiqui \textit{et al.} \cite{siddiqui2019rethinking}. Just like Schreiber \textit{et al.} \cite{schreiber2017deepdesrt}, they have formulated the problem of structure recognition as the semantic segmentation problem. The authors have used fully convolutional networks \cite{long2015fully} to segment the rows and columns respectively. Assuming the consistency in a tabular structure, the method of prediction tiling is introduced which reduces the complexity of table structural recognition. The author used the structural models of FCN's encoder and decoder, and loaded pre-trained models on ImageNet \cite{russakovsky2015imagenet}. Given an image, the model produces the features having the same size as the original input image. The tiling process averages the features in rows and columns and combines the features of \textit{$H \times W \times C$} ($Height \times Width \times Channel$) into \textit{$H \times C$} for rows and \textit{$W \times C$} for columns. Features after being convolved are expanded into \textit{$H \times W \times C$}. Subsequently, the label of each pixel is obtained through the convolution layer. Finally, post-processing is performed to accomplish the final result. The authors have reported the F1-score of 93.42\% with an IOU of 0.5 on the ICDAR 2013 dataset \cite{icdar13}. Due to the writer's constraint of consistency, they have to finetune this dataset which is now publicly available to reproduce similar results\footnote{Fine-tuned ICDAR-13 dataset : https://bit.ly/2NhZHCr}.

\begin{figure}
    \begin{tikzpicture}
        \begin{axis}[
            title=\textbf{Methods used 
            in Table Structure Segmentation },
            ybar,
            enlargelimits=0.15,
            legend style={at={(0.5,-0.15)},
              anchor=north,legend columns=-1},
            ylabel={Count},
            symbolic x coords={OD, SIS,RNN,DDC,GNN},
            xtick=data,
            nodes near coords,
            nodes near coords align={vertical},
            ]
        \addplot coordinates {(OD,4) (SIS,3) (RNN,2) (GNN,3) (DDC,2)};
        %\legend{Deep Learning Concepts}
        \end{axis}
    \end{tikzpicture}
\captionof{figure}{OD denotes Object Detection, SIS is Semantic Image Segmentation, RNN represents Recurrent Neural Networks, DDC is an abbreviation for Deformable and Dilated Convolutions whereas GNN is Graphical Neural Networks. This graph explains that what kind of deep learning algorithms are periodically exploited to perform table structure segmentation.}
\label{graph:segmentation}
\end{figure}
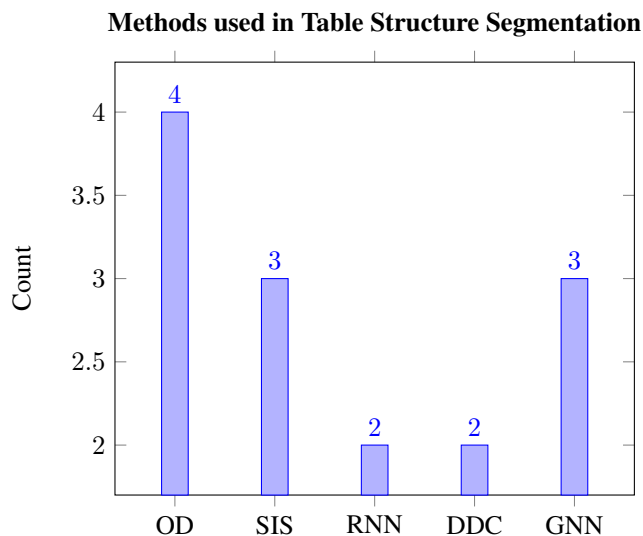

Zou and Ma \cite{zou2020deep} proposed another research in which  fully convolutional networks \cite{long2015fully} are utilized to develop image-based table structure recognition method. Similar to the idea of \cite{siddiqui2019rethinking}, the presented work segments the rows, columns, and cells in a table. Connected Component Analysis is used to improve the predicted boundaries of all of the table components \cite{dillencourt1992general}. Later, row and column numbers are assigned for each cell based on the position of row and column separators. Moreover, custom heuristics are applied to optimize cellular boundaries.

\begin{subsubsection}{Graph Neural Networks}
\label{sec:gnn_seg}
So far in most of the mentioned approaches, the problem of segmenting tables in document images is treated with segmentation techniques. In 2019, Qasim \textit{et al.} \cite{qasim2019rethinking} exploited the graph neural networks \cite{scarselli2008graph} to perform table recognition for the first time. The model is constructed with a blend of deep convolutional neural networks to extract image features and graph neural networks to control the relationship among the vertices. They have open-sourced the proposed work to reproduce or improve the claimed results. \footnote{github.com/shahrukhqasim/TIES-2.0}

Another technique powered by graph neural networks to recognize the tabular structure is proposed in the same year by \textit{Chi et al.} \cite{chi2019complicated}. However, this technique is based on PDF documents instead of images. One contribution from their side worth mentioning is the publication of their large-scale table structure recognition dataset \textit{SciTSR} which will be discussed in Section \ref{sec:datasets}.
\end{subsubsection}

\paragraph{Distance Based Weights}
\label{sec:dbw}
Another work to segment tabular structures presented in ICDAR 2019 is about the reconstruction of syntactic structures from the table as known as \textit{ReS${^2}$TIM} published by Xue \textit{et al.} \cite{xue2019res2tim}. The primary goal of this model is to regress the coordinates for each cell. The novel approach first creates a network that detects neighbors of each cell in a table. Distance-based weight is presented in the paper which will help the network to solve the class imbalance hurdle during training. Experiments were carried out on Chinese medical documents dataset \cite{xue2018table} and ICDAR 2013 table competition dataset \cite{icdar13}.

\begin{table*}
\centering

\caption{A summary of advantages and limitations of various deep learning-based methods that have worked on the task of table structure recognition. The bold horizontal line separates the approaches with different architectures.}

\normalsize
\setlength\extrarowheight{5pt}
\newcolumntype{l}{>{\hsize=.3\hsize}X}
\newcolumntype{m}{>{\hsize=.5\hsize}X}
\newcolumntype{a}{>{\hsize=.5\hsize}X}
\newcolumntype{b}{>{\hsize=.5\hsize}X}

\begin{tabularx}{\textwidth}{>{\justifying\arraybackslash\noindent}l|m|a|b}

    \specialrule{.2em}{.1em}{.1em} 
    \captionsetup{font=scriptsize}
    \textbf{Literature}&
    \textbf{Method}&
    \textbf{Highlights}&
    \textbf{Limitations}\\

    \specialrule{.2em}{.1em}{.1em}
    
    \footnotesize \textit{Siddiqui et al.} \cite{siddiqui2019deeptabstr}& 
    \footnotesize Deformable Convolution with Faster R-CNN (Section \ref{sec:ddc}).&
    \footnotesize \textbf{a)} Published another dataset having structural information of tables. \textbf{b)} Deformable convolution allows tackling varied tabular structures. & 
    \footnotesize The published work will not work well in case of row/column span in the tables. \\
    
    \hline
   
    \footnotesize \textit{CascadeTabNet} \cite{prasad2020cascadetabnet}& 
    \footnotesize Cascade Mask R-CNN with HRNet as a backbone network (Section \ref{sec:odstr}).&
    \footnotesize An end-to-end approach to directly regress cellular boundaries.& 
    \footnotesize An extra post-processing is required to filter tables (with and without) ruling lines.\\

    \hline
    
    \footnotesize \textit{GTE} \cite{zheng2021global}& 
    \footnotesize Generic object detection approach (Section \ref{sec:odstr}).&
    \footnotesize An hierarchical network with an additional novel cluster-based method to recognize tabular structures. & 
    \footnotesize Final cell structure recognition is conditioned on the precise classification of a table (Graphical ruling lines present or not present).\\
    \hline
    
    \footnotesize \textit{Hashmi et al.} \cite{hashmi2021guided}& 
    \footnotesize Mask R-CNN with an Anchor optimization method (Section \ref{sec:odstr}).&
    \footnotesize Optimized anchors help region proposal networks to converge faster and better.& 
    \footnotesize This work depends on the initial pre-processing step of clustering the ground-truth to retrieve suitable anchors.\\
    \hline

    \footnotesize \textit{Raja et al.} \cite{raja2020table}& 
    \footnotesize  Mask R-CNN with ResNet-101 as a backbone network (Section \ref{sec:odstr}).&
    \footnotesize \textbf{a)} A trainable combination of top-down (cell detection) and bottom-up (structure recognition) is presented. \textbf{b)} An additional alignment loss is proposed to detect cells accurately. & 
    \footnotesize The approach is vulnerable in the case of empty cells.\\
    
    \specialrule{.2em}{.1em}{.1em}
    
    \footnotesize \textit{Siddiqui et al.} \cite{siddiqui2019rethinking} &
    \footnotesize
    Fully Convolutional Networks (Section \ref{sec:fcn_seg}). & 
    \footnotesize The proposed Prediction tiling technique minimizes the complexity of the problem of table structure recognition. & 
    \footnotesize \textbf{a)} The method relies on the consistency assumption of tabular structures, \textbf{b)} In case of overly-segmented rows/columns, extra post-processing steps are required. \\
    
    \hline
    
    \footnotesize \textit{Tensmeyer et al.} \cite{tensmeyer2019deep}& 
    \footnotesize Dilated Convolutions in Fully Convolutional Networks (Section \ref{sec:dddc}). &
    \footnotesize The system works well both on PDF and scanned document images. & 
    \footnotesize The merging part of the approach is depends on the post-processing heuristics. \\
    \hline
    
    \footnotesize \textit{Zou et al.} \cite{zou2020deep}& 
    \footnotesize Fully Convolutional Networks (Section \ref{sec:fcn_seg}). & 
    \footnotesize \textbf{a)} Along with segmenting rows and columns, cells are segmented in a table. \textbf{b)} Applying Connected component analysis further improves the results.& 
    \footnotesize Handful of post-processing steps involving custom heuristics are required to produce comparative results.\\
    \specialrule{.2em}{.1em}{.1em}
    
    \footnotesize \textit{Qasim et al.} \cite{qasim2019rethinking}& 
    \footnotesize Graph Neural Networks with Convolutional Neural Networks (Section \ref{sec:gnn_seg}).&
    \footnotesize \textbf{a)} The proposed method exploits both the spatial and textual features, \textbf{b)} A novel Monte Carlo based memory efficient training method is also presented in this work. & 
    \footnotesize The system is not evaluated on the publicly available table datasets. \\
    
    \hline
    
    \footnotesize \textit{Xue et al.} \cite{xue2019res2tim}& 
    \footnotesize Graph Neural Networks with distance based weights (Section \ref{sec:dbw}).&
    \footnotesize The distance-based weight technique resolves the class imbalance problem for the cell relationship network.& 
    \footnotesize  The method is vulnerable in the case of sparse tables.\\
        
    \specialrule{.2em}{.1em}{.1em}
    
    \footnotesize \textit{Khan et al.} \cite{khan2019table}& 
    \footnotesize Recurrent Neural Networks (Section \ref{sec:rnn}).&
    \footnotesize The Bi-directional GRU overcomes the problem of the smaller receptive field of CNNs.& 
    \footnotesize A series of pre-processing steps such as binarization, noise removal, and morphological transformation are required. \\
    
    \specialrule{.2em}{.1em}{.1em}

\end{tabularx} 
\label{tab:comparison-segmentation}
\end{table*}

\subsubsection{Recurrent Neural Networks}
\label{sec:rnn}
So far, we have seen that convolutional neural networks and graph neural networks are employed to perform table structure extraction. Recent research proposed by Khan \textit{et al.} \cite{khan2019table} has experimented with bi-directional recurrent neural networks along with Gated Recurrent Units (GRU) \cite{chung2014empirical} to extract the structure of the table. The authors argue that the receptive field of the convolutional neural network is not capable enough to capture complete information of row and column in one stride. According to the writers, a pair of bi-directional GRU performs better. One GRU caters to the row identification whereas another detects the column boundary. The author tried two classic recurrent neural network models, Long Short Term Memory (LSTM) \cite{hochreiter1997long} and GRU \cite{chung2014empirical}, and found that GRU has more benefits in experimental results. In the end, the authors experimented with the datasets of the table structure recognition sub-task of the UNLV \cite{shahab2010open} and ICDAR 2013 table competitions, both surpassing the previous best results. The authors tried to convince that GRU-based sequential models can also be exploited to improve not only the problem of structure recognition but also for the information extraction in the tables.

Besides the huge dataset, the author of \textit{TableBank} \cite{li2020tablebank} has published the baseline model for the table structure recognition. Image-to-markup model \cite{deng2016you} is trained on \textit{TableBank} dataset. To implement the model, OpenNMT \cite{klein2017opennmt} is applied which is an open source tool kit for neural machine translation.

\subsubsection{Deformable and Dilated Convolutions}
\label{sec:ddc}
Along with traditional convolutions, deformable and dilated convolutions have been exploited to recognize tabular structures in document images.
\begin{paragraph}{Deformable Convolutions}
Siddiqui \textit{et al.} \cite{siddiqui2019deeptabstr} advertised another public image-based table recognition dataset known as \textit{TabStructDB}. This dataset was curated by using the images from a well known ICDAR 2017 page object detection dataset \cite{icdar17} which are annotated with structural information. \textit{TabStructDB} has been extensively evaluated on the proposed model called \textit{DeepTabStR} which can be seen as a follow-up work for \cite{siddiqui2018decnt}. The author stated that there exists a huge diversity in the tabular layouts and traditional convolutions which operates as a sliding window is not the best choice. Deformable convolutions \cite{dai2017deformable} allows the network to adjust the receptive field by considering the current position of an object. Hence, the author leverages the deformable convolution to perform the task of structural recognition of tables. The exercise of table segmentation is operated as an object detection problem in this research. Deformable Faster R-CNN is used in \textit{DeepTabStR}, where the traditional ROI-pooling layer is replaced with a deformable ROI-pooling layer. Another important point is highlighted in this research that there still exists room for improvement in the area of structural analysis of tables having inconsistent layouts.

\end{paragraph}

\begin{paragraph}{Dilated Convolutions}
\label{sec:dddc}
Another technique employing dilated convolutions \textit{SPLERGE} (Split and Merge models) is proposed by Tensmeyer \textit{et al.} \cite{tensmeyer2019deep}. Their approach consists of two separate deep learning models in which the first model defines the grid-like structure of the table whereas the second model finds out whether cells can be further spanned into multiple rows or columns. The author claims to achieve state-of-the-art performance on the ICDAR 2013 table competition dataset \cite{icdar13}.

\end{paragraph}

\subsubsection{Object Detection Algorithms}
\label{sec:odstr}

Inspiring from the exceptional results of object detection algorithms \cite{he2017mask, cai2018cascade}, researchers in the table community have formulated the task of table structure recognition as an object detection problem. 

Hashmi \textit{et al.} \cite{hashmi2021guided} proposed a guided table structure recognition method to detect rows and columns in tables. This paper presents that the localization of rows and columns can be improved by incorporating an anchor optimization method \cite{wang2019region}. In their proposed work, Mask R-CNN \cite{he2017mask} is employed with optimized anchors to detect the boundaries of rows and columns. The presented work has reported state-of-the-art results on TabStructDB \cite{siddiqui2019deeptabstr} and table structure recognition dataset of ICDAR-2013 (released by \cite{schreiber2017deepdesrt}).

Until now, we have discussed approaches that detect tabular rows and columns to retrieve the final structure of a table. Contrary to the previous approaches, Raja et al. \cite{raja2020table} introduced a table structure recognition method that directly regresses the cellular boundaries. The authors employed Mask R-CNN \cite{he2017mask} with a ResNet-101 backbone pre-trained on MS-COCO dataset \cite{lin2014microsoft}. In their object detection framework, dilated convolutions \cite{yu2015multi} are implemented in the region proposal network. Furthermore, the authors introduced alignment loss that also contributes to the overall loss function. Later, graph convolutional networks \cite{kipf2016semi} are applied to obtain the row and column relationship between the predicted cells. The whole process is trained in an end-to-end fashion. The paper presents extensive evaluations on several publicly available datasets for the task of table structure recognition.

Another approach that directly localize the cellular boundaries in tables is presented in \textit{CascadeTabNet} \cite{prasad2020cascadetabnet}. In this approach, tabular images are given to the Cascade Mask R-CNN \cite{cai2018cascade} that predicts the cellular mask along with the classification of the table as bordered or borderless. Subsequently, individual post-processing is applied to bordered and borderless tables to retrieve the final cellular boundaries.

The system GTE proposed by Zheng \textit{et al.} \cite{zheng2021global} is an end-to-end framework that not only detects the tables but recognizes the structures of tables in document images. Analogous to the approach of \cite{prasad2020cascadetabnet}, the authors have suggested two different cell detection networks i.e: 1) For graphical ruling lines present in a table. 2)  No graphical ruling lines in a table. Instead of a tabular image, a complete document image with a table mask is propagated to the classification network. Based on the predicted class, the image is passed to the appropriate cell network to retrieve the final cell boundaries.

\begin{table*}
\centering

\caption{A summary of advantages and limitations of deep learning-based methods that have solely worked on the task of table recognition on scanned document images.}
\normalsize
\setlength\extrarowheight{5pt}
\newcolumntype{l}{>{\hsize=.3\hsize}X}
\newcolumntype{m}{>{\hsize=.5\hsize}X}
\newcolumntype{a}{>{\hsize=.5\hsize}X}
\newcolumntype{b}{>{\hsize=.5\hsize}X}

\begin{tabularx}{\textwidth}{>{\justifying\arraybackslash\noindent}l|m|a|b}

    \specialrule{.2em}{.1em}{.1em} 
    \captionsetup{font=scriptsize}
    \textbf{Literature}&
    \textbf{Method}&
    \textbf{Highlights}&
    \textbf{Limitations}\\

    \specialrule{.2em}{.1em}{.1em}
    
    \footnotesize \textit{Zhong et al.} \cite{zhong2019image} &
    \footnotesize
    Attention based encoder dual decoder (Section \ref{sec:enc-dual-dec}). & 
    \footnotesize \textbf{a)} Published a large-scale table dataset, \textbf{b)} The approach presents a novel evaluation metrics \textit{TEDS} to evaluate table recognition methods. & 
    \footnotesize The approach is not directly comparable with other state-of-the-art methods. \\
    
    \hline
    
    \footnotesize \textit{Deng et al.} \cite{deng2019challenges}& 
    \footnotesize Encoder decoder network presented as the baseline model (Section \ref{sec:enc-dec}).&
    \footnotesize \textbf{a)} Contributed with another large-scale dataset in the field of table understanding, \textbf{b)} Challenges in end-to-end table recognition are discussed in the presented work. & 
    \footnotesize The proposed baseline method is not evaluated on the other publicly available table recognition datasets. \\
    \specialrule{.2em}{.1em}{.1em}

\end{tabularx} 
\label{tab:comparison-recognition}
\end{table*}

\subsection{Table Recognition}
\label{sec:table_rec}
As explained in Section \ref{sec:methodology}, the task of table recognition covers the job of table structure extraction along with extracting the text from the table cells. Relatively, less progress has been accomplished in this specific domain.

In this section, we will cover the recent experiments that have attempted the problem of table recognition. Table \ref{tab:comparison-recognition} summarizes these approaches by highlighting their advantages and limitations.

\subsubsection{Encoder-Dual-Decoder}
\label{sec:enc-dual-dec}
Recently, research on image-based table recognition proposed by Zhong \textit{et al.} \cite{zhong2019image} is published. In this research, the authors proposed a new dataset known as \textit{PubTabNet} which is explained in Section \ref{sec:pubtabnet}. The authors have attempted to resolve the problem of inferring both the structure recognition of tables and the information present in their respective cells. The writers of the paper have treated the task of structure recognition and table recognition separately. They proposed the attention-based Encoder-Dual-Decoder (EDD) architecture. The encoder extracts the essential spatial features, then the first decoder segments the table into rows and columns whereas another decoder attempts to identify the content of a table cell. In this research, a new Tree-Edit-Distance-based Similarity (TEDS) metrics is presented to evaluate the quality of cell content identification.

\subsubsection{Encoder Decoder Network}
\label{sec:enc-dec}
Another dataset \textit{TABLE2LATEX-450K}\footnote{https://github.com/bloomberg/TABLE2LATEX.} has been published recently in the ICDAR conference comprises of arXiv articles. Along with the dataset, Deng \textit{et al.} \cite{deng2019challenges} discussed the current challenges in the end-to-end table recognition and highlights the worth of a bigger dataset in this field. The creators of this dataset have also conferred the baseline models ( \textit{IM2TEX}) \cite{deng2017image} on the mentioned dataset by using an encoder-decoder architecture with an attention mechanism. IM2TEX model is implemented on OpenNMT \cite{klein2017opennmt}. With the probable increase in hardware capabilities of the GPUs in the future, the authors claim that this dataset will be proved as a promising contribution.

It is important to mention that apart from these two approaches, other methods \cite{li2020tablebank, qasim2019rethinking, rashid2017table} have extracted the contents of cells in order to recognize either the tabular boundaries or tabular structures. 

\section{Datasets}
\label{sec:datasets}

\begin{figure*}[ht]
    \includegraphics[width = \linewidth]{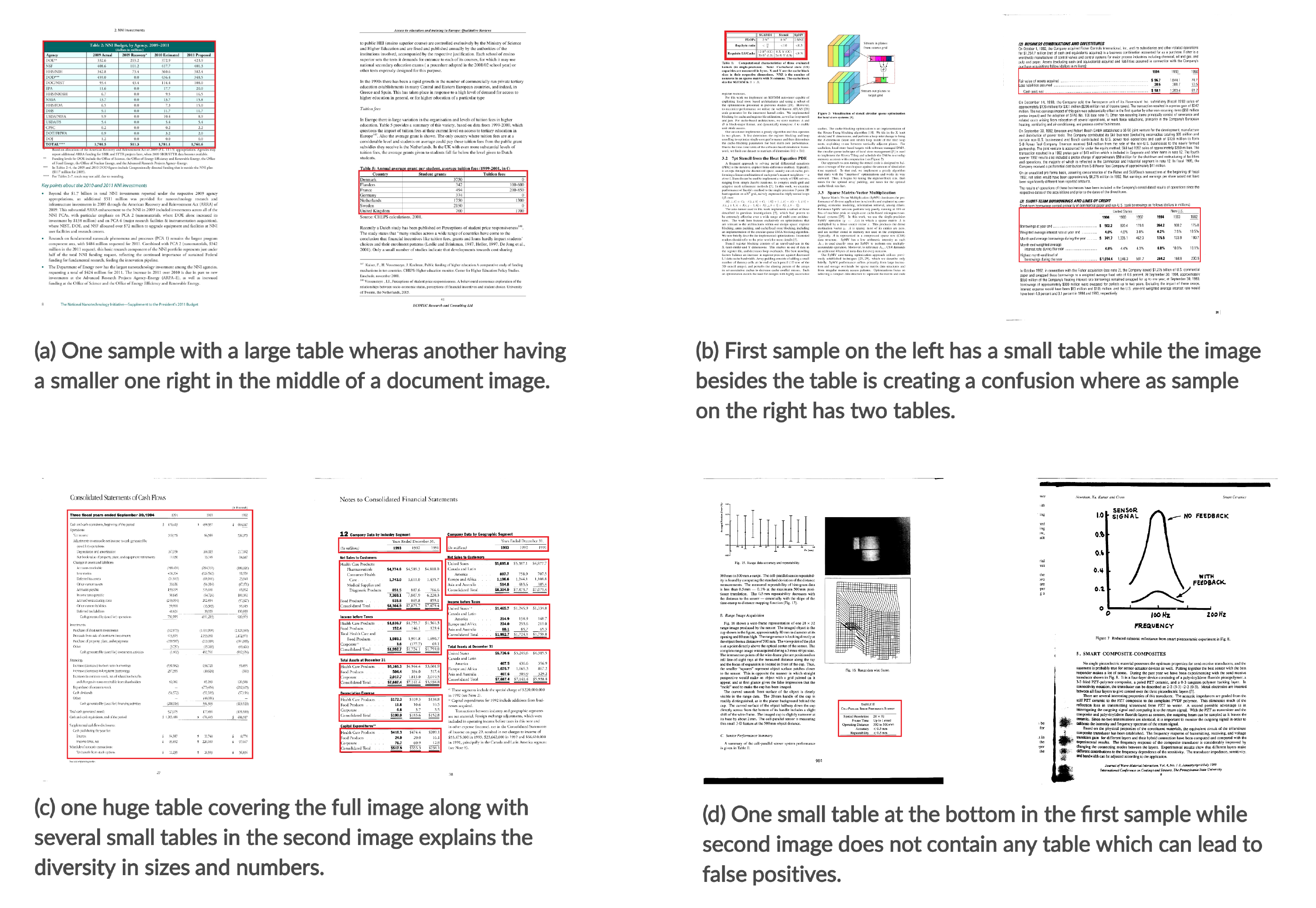}
    \caption{Sample document images taken from the datasets of ICDAR-2013 \cite{icdar13}, ICDAR-2017-POD \cite{icdar17}, UNLV \cite{shahab2010open} and UW3 \cite{phillips1996user}. The red boundaries represent the tabular region. The diversity between samples in a dataset is quite evident.  }
    \label{fig:dataset}
\end{figure*}

The performance of deep neural networks has a direct relation with the size of the dataset \cite{schreiber2017deepdesrt,siddiqui2018decnt}. In this section, we will discuss all of the well-known datasets that are publicly available to deal with the problem of table detection and table structural recognition in document images.
Table \ref{tab:dataset} contains a comprehensive explanation of all the mentioned datasets which are employed to perform and compare detection, structural segmentation and recognition of tables in document images. Figure \ref{fig:dataset} demonstrates samples from some of the distinguished datasets in table community.

\subsection{ICDAR-2013}
\label{sec:icdar_13}
International Conference on Document Analysis and Recognition (ICDAR) 2013 \cite{icdar13} is the most renowned dataset among the researchers in the table community. This dataset is published for the table competition organized by the ICDAR conference in 2013. This dataset has the annotations for both table detection and table recognition. The dataset consists of PDF files which are often converted into images to be utilized in the various approaches. The dataset contains structured tables, graphs, charts, and text as information. There are a total of 238 images in the dataset, out of which 128 incorporates tables. This dataset has been extensively used to compare state-of-the-art approaches. As mentioned in the Table \ref{tab:dataset}, this dataset has annotations for all of the three tasks of table understanding which are discussed in the paper. A couple of samples from this dataset are illustrated in Figure \ref{fig:dataset} (a).

\subsection{ICDAR-2017-POD}
\label{sec:icdar_17}
This dataset \cite{icdar17} is also proposed for the competition of Page Object Detection (POD) in ICDAR 2017. This dataset is widely used to evaluate approaches for table detection. This dataset is fairly bigger than the ICDAR 2013 table dataset. It comprises of total 2417 images including tables, formulas, and figures. In many instances, this dataset is divided into 1600 images (731 tabular regions) which are used for training while the rest of 817 images (350 tabular regions) are employed for the testing purpose. A pair of instances of this dataset are demonstrated in Figure \ref{fig:dataset} (b). This dataset has only information for the tabular boundaries as explained in Table \ref{tab:dataset}.

\subsection{UNLV}
\label{sec:unlv}
The UNLV dataset \cite{shahab2010open} is a recognized dataset in the field of document image analysis. This dataset composed of scanned document images from various sources like financial reports, magazines, and research papers having diverse tabular layouts. Although the dataset contains approximately 10,000 images, only 427 images contain tabular regions. Frequently, these 427 images have been used to conduct various experiments in the research community. This dataset has been used for all the three tasks of table analysis which are discussed in the paper. Figure \ref{fig:dataset} (c) illustrates a couple of samples from this dataset.

\subsection{UW3}
\label{sec:uw3}
UW3 \cite{phillips1996user} is another popular dataset for researchers working in the area of document image analysis. This dataset contains scanned documents from books and magazines. There are approximately 1600 scanned document images out of which only 165 images have table regions. Annotated table coordinates are present in the XML format. Two samples from this dataset are demonstrated in Figure \ref{fig:dataset} (d). Although this dataset has limited number of tabular regions, it has annotations for all the three problems of table understanding that are discussed in the paper.

\subsection{ICDAR-2019}
\label{sec:icdar_19}
\begin{figure}[ht]
    \includegraphics[width = \columnwidth]{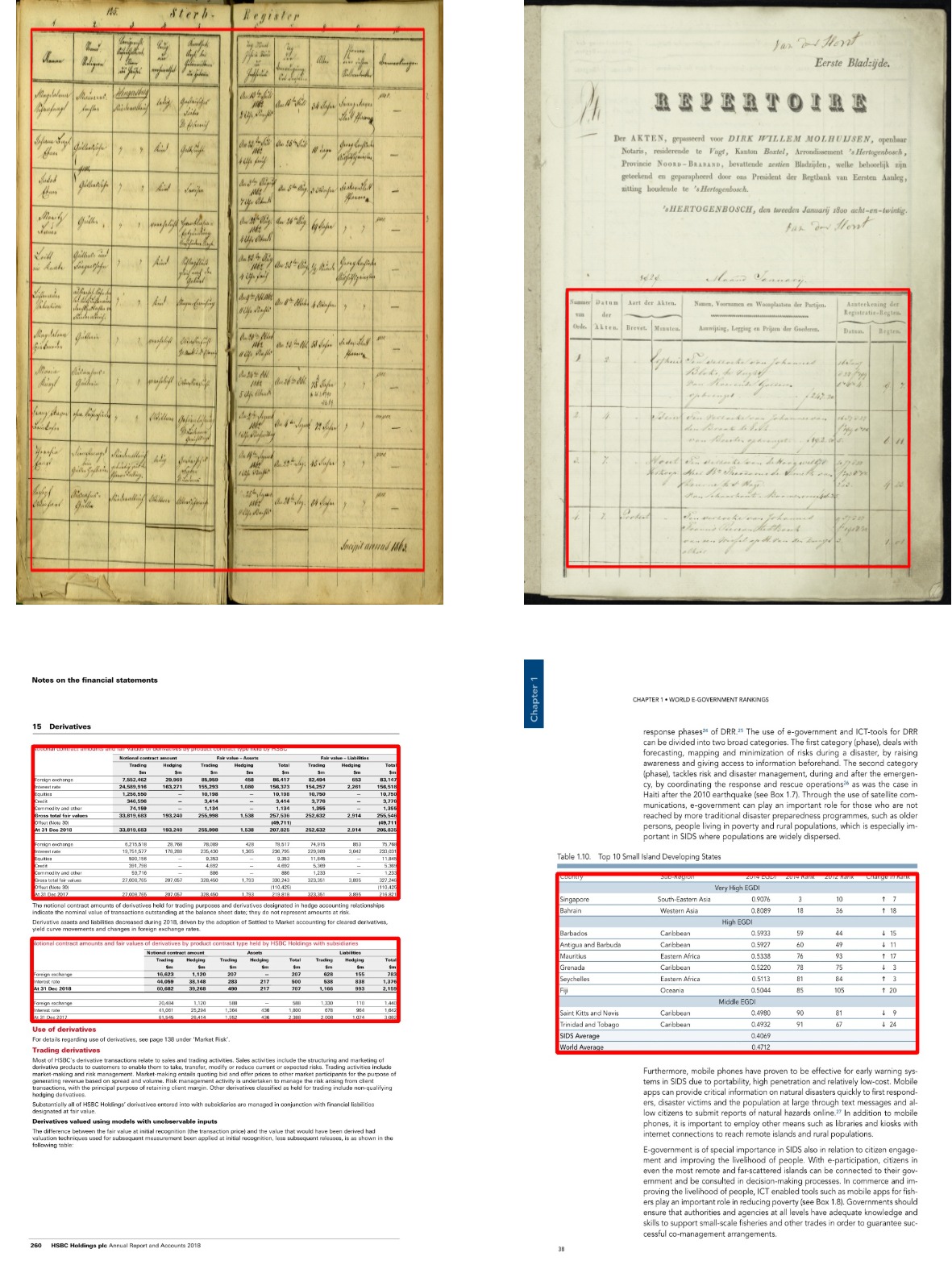}
    \caption{Examples of archival and modern document images taken from the ICDAR-2019 dataset \cite{icdar19} which is explained in Section \ref{sec:icdar_19}. The red boundaries represent the tabular region.}
    \label{fig:icdar_19}
\end{figure}

Recently, Competition on Table Detection and Recognition (cTDaR) \cite{icdar19} is carried out in ICDAR 2019. In the competition, two new datasets are proposed: modern and historical datasets. The modern dataset contains samples from scientific papers, forms, and financial documents. Whereas the archival dataset includes images from hand-written accounting ledgers, schedules of train, simple tabular prints from old books, and many more. The prescribed train-test split for detecting tables in the modern dataset is 600 images for training while 240 images for the test. Similarly, for the historical dataset 600 images for the training and 199 images for the testing part are the recommended data distribution. As summarized in Table \ref{tab:dataset}, the dataset has information for tabular boundaries and annotations for the cell area as well. This novel dataset is challenging in nature because it contains both modern and historical (archived) document images. This dataset will be used to evaluate the robustness of table analysis methods. In order to understand the diversity, a couple of samples from both the historical and modern datasets are depicted in Figure \ref{fig:icdar_19}.

\subsection{Marmot}
\label{sec:marmot}
Not long ago, Marmot \footnote{http://www.icst.pku.edu.cn/cpdp/sjzy/index.htm} is one of the largest publicly available datasets and extensively used by the researchers in the area of table understanding. This dataset has been proposed by the Institute of Computer Science and Technology (Peking University) and later explained by Fang \textit{et al.} \cite{fang2012dataset}. There are 2000 images in the dataset composed of English and Chinese conference papers from 1970 to 2011. The dataset is highly useful for training the networks due to having diverse and very complex page layouts. There is a roughly 1:1 ratio between positive to negative images in the dataset. Some occasions of incorrect ground-truth annotations have been reported in the past which are later cleaned by Schreiber \textit{\textit{et al.}} \cite{schreiber2017deepdesrt}. As mentioned in Table \ref{tab:dataset}, this dataset has annotations for the tabular boundaries and it is widely exploited to train deep neural networks for table detection.

\subsection{TableBank}
\label{sec:TableBank}
In early 2019, Minghao \textit{et al.} \cite{li2020tablebank} realized the need for large datasets in the table community and published TableBank, a dataset comprising of 417 thousand labeled images having tabular information. This dataset has been collected by crawling over documents available online in \textit{.docx} format. Another source of data for this dataset is LaTeX documents which were collected from the database of arXiv \footnote{https://arxiv.org}. The publishers of this dataset argue that this contribution will facilitate the researchers to leverage the power of deep learning and fine-tuning methods. The authors claim that this dataset can be used for both table detection and structural recognition tasks. However, we are unable to find annotations for structural recognition in the dataset \footnote{https://github.com/doc-analysis/TableBank}. Important information for the dataset is summarized in Table \ref{tab:dataset}.

\subsection{TabStructDB}
\label{sec:tabStruct}
In the ICDAR conference 2019, along with the table competition \cite{icdar19}, other researchers have also published new datasets in the field of table analysis. One of the dataset known as  TabStructDB\footnote{https://bit.ly/2XonOEx} is published by Siddiqui \textit{et al.} \cite{siddiqui2019deeptabstr}. Since the ICDAR-2017-POD dataset \cite{icdar17} has only information for the tabular boundaries, the author leverages this dataset and annotated them with structural information comprising of boundaries of respective rows and columns in the table. To maintain consistency, the authors have also kept the same dataset split as mentioned in \cite{icdar19}. Significant information regarding the dataset is summarized in Table \ref{tab:dataset}. Since this dataset provides information regarding the boundaries of rows and columns, it facilitates the researchers to treat the task of table structure recognition as object detection or semantic segmentation problem.

\subsection{TABLE2LATEX-450K}
\label{sec:tab2latex}
Another large dataset that is published in the recent ICDAR conference is TABLE2LATEX-450K \cite{deng2019challenges}. The dataset contains 450 thousand annotated tables along with their corresponding images. This huge dataset is constructed by crawling over the arXiv articles from the year 1991 to 2016 and all the LaTeX source documents were downloaded. After the extraction of source code and subsequent refinement, the high-quality labeled dataset is obtained. As mentioned in Table \ref{tab:dataset}, the dataset contains annotations for the structural segmentation of tables and the content of table cells. Along with the dataset, publishers have made all the pre-processing scripts publicly available\footnote{https://github.com/bloomberg/TABLE2LATEX.}. This dataset is an important contribution to tackle the problem of table structural segmentation and table recognition in document images because it enables the researchers to train the massive deep learning architectures from scratch which can be further fine-tuned on relatively smaller datasets.

\newcommand{\cmark}{\ding{51}}%
\newcommand{\xmark}{\ding{55}}%
\newcolumntype{d}{X}
\newcolumntype{b}{>{\hsize=.9\hsize}X}
\newcolumntype{s}{>{\hsize=.15\hsize}X}
\newcolumntype{o}{>{\hsize=.3\hsize}X}

\begin{table*}[ht]
\centering

\caption{Table Datasets. TD denotes Table Detection, TSR is Table Structure Recognition wheras TR is Table Recognition.}
\normalsize
\setlength\extrarowheight{5pt}
% default value: 6pt{p{0.1\textwidth}p{0.8\textwidth}}
% \begin{tabularx}{\textwidth}{|C|C|C|C|C|C|C|} 
\begin{tabularx}{\textwidth}{b|s|s|s|o|o|d}
\specialrule{.2em}{.1em}{.1em} 
\captionsetup{font=scriptsize}
\textbf{Dataset}&
\textbf{TD}&
\textbf{TSR}&
\textbf{TR}&
\textbf{\# Samples}&
\textbf{Image Type}&
\textbf{Location}\\
\specialrule{.2em}{.1em}{.1em} 
% \hline
\footnotesize
{ICDAR-2013} \cite{icdar13} (Section \ref{sec:icdar_13}) &
\cmark&
% \verb|\xmark|: \xmark&
\cmark&
\cmark&
\small
{238}&
\footnotesize{Scanned}&
\small
\footnotesize{\href{http://www.tamirhassan.com/html/dataset.html}{http://www.tamirhassan.com/html/dataset.html}}\\
\hline

\footnotesize {ICDAR-2017-POD\cite{icdar17}} (Section \ref{sec:icdar_17}) &
\cmark&
\xmark&
\xmark&
\small
{2.4K}&
\footnotesize{Scanned}&
\small
\footnotesize{\href{http://www.icst.pku.edu.cn/cpdp}{http://www.icst.pku.edu.cn/cpdp}}\\
\hline
\footnotesize
{ICDAR-2019}\cite{icdar19} (Section \ref{sec:icdar_19})&
\cmark&
\cmark&
\xmark&
\small
{3.6K}&
\footnotesize{Scanned}&
\footnotesize{\href{https://zenodo.org/record/2649217}{https://zenodo.org/record/2649217}}\\
\hline
\footnotesize
{UNLV \cite{shahab2010open} (Section \ref{sec:unlv})}&
\cmark&
\cmark&
\cmark&
\small
427&
\footnotesize{Scanned}&
\footnotesize{\href{https://drive.google.com/file/d/1ETq5hhoIgCzzom6yivkokhQ8DoOm6nDs}{https://drive.google.com/file/d/}}\\
\hline
\footnotesize
{Marmot \cite{fang2012dataset} (Section \ref{sec:marmot})} &
\cmark&
\xmark&
\xmark&
\small
{958}&
\footnotesize{Scanned}&
\footnotesize{\href{http://www.icst.pku.edu.cn/cpdp/sjzy/}{http://www.icst.pku.edu.cn/cpdp/sjzy/}}\\
\hline
\footnotesize
{UW3 \cite{phillips1996user} (Section \ref{sec:uw3})}&
\cmark&
\cmark&
\cmark&
\small
\small{165}&
\footnotesize{Scanned}&
\footnotesize{\href{http://tc11.cvc.uab.es/datasets/DFKI\-TGT\-2010\_1/}{http://tc11.cvc.uab.es/datasets/DFKI\-TGT\-2010\_1/}}\\
\hline
\footnotesize
{TableBank \cite{li2020tablebank} (Section \ref{sec:TableBank})}&
\cmark&
\cmark&
\xmark&
\footnotesize{ $417$K$ (TD)$, $145$K (TSR)}&
\footnotesize{Scanned}&
\footnotesize{\href{https://github.com/doc-analysis/TableBank/}{https://github.com/doc-analysis/TableBank}}\\
\hline
\small
\footnotesize{TabStructDB \cite{siddiqui2019deeptabstr} (Section \ref{sec:tabStruct})}&
\xmark&
\cmark&
\xmark&
\small
\small{2.4K}&
\footnotesize{Scanned}&
\footnotesize{\href{https://bit.ly/2XonOEx}{https://bit.ly/2XonOEx}}\\
\hline

\footnotesize{TABLE2LATEX \cite{deng2019challenges} (Section \ref{sec:tab2latex})}&
\xmark&
\cmark&
\cmark&
\small
{450K}&
\footnotesize{Scanned}&
\footnotesize{\href{https://github.com/bloomberg/TABLE2LATEX/}{https://github.com/bloomberg/TABLE2LATEX}}\\
\hline
\footnotesize
{SciTSR \cite{chi2019complicated} (Section \ref{sec:scitsr})}&
\xmark&
\cmark&
\cmark&
\small
\small{15K}&
\footnotesize{Scanned}&
\footnotesize{\href{https://github.com/Academic-Hammer/SciTSR}{https://github.com/Academic-Hammer/SciTSR}}\\
\hline
\footnotesize
{DeepFigures \cite{kim2008extracting} (Section \ref{sec:deepfigure})}&
\cmark&
\xmark&
\xmark&
\small
{1.4M}&
\footnotesize{Scanned}&
\footnotesize{\href{https://s3-us-west-2.amazonaws.com/ai2-s2-research-public/deepfigures/jcdl-deepfigures-labels.tar.gz}{https://s3-us-west-2.amazonaws.com/ai2-s2-research-public/}}\\
\hline
\footnotesize
{RVL-CDIP (Subset) (Section \ref{sec:rvlcdip})} \cite{riba2019table}&
\cmark&
\xmark&
\xmark&
\small
{518}&
\footnotesize{Scanned}&
\footnotesize{\href{https://zenodo.org/record/3257319}{https://zenodo.org/record/3257319}}\\
\hline
\footnotesize
{PubTabNet \cite{zhong2019image} (Section \ref{sec:pubtabnet})}&
\xmark&
\cmark&
\cmark&
\small
{568K}&
\footnotesize{Scanned}&
\footnotesize{\href{https://github.com/ibm-aur-nlp/PubTabNet}{https://github.com/ibm-aur-nlp/PubTabNet}}\\
\hline
\footnotesize
{IIT-AR-13k \cite{mondal2020iiit} (Section \ref{sec:iitar})}&
\cmark&
\xmark&
\xmark&
\small
{13K}&
\footnotesize{Scanned}&
\footnotesize{\href{http://cvit.iiit.ac.in/usodi/iiitar13k.php}{http://cvit.iiit.ac.in/usodi/iiitar13k.php}}\\
\hline
\footnotesize
{CamCap \cite{seo2015junction} (Section \ref{sec:CamCap})}&
\cmark&
\cmark&
\xmark&
\small
{75}&
\footnotesize{Camera-captured}&
\footnotesize{\href{http://ispl.snu.ac.kr/~cusisi/dataset.zip}{http://ispl.snu.ac.kr/~cusisi/dataset.zip}}\\
\specialrule{.2em}{.1em}{.1em} 
\end{tabularx}
\label{tab:dataset}
\end{table*}

\subsection{SciTSR}
\label{sec:scitsr}
SciTSR is another dataset released in 2019 by Zewen, \textit{et al.} \cite{chi2019complicated}. According to the authors, this is one of the largest publicly available dataset for the task of table structure recognition \footnote{https://github.com/Academic-Hammer/SciTSR}. The dataset consists of 15 thousands tables in PDF format along with its annotations. The dataset is constructed by crawling LaTeX source files from the arXiv. Roughly 25\% of the dataset consists of complicated tables that span into multiple rows or columns. This dataset has annotations for table structural segmentation and table recognition as summarized in Table \ref{tab:dataset}. Because of having complex tabular structures, this dataset can be exploited to improve state-of-the-art systems dealing with structural segmentation and recognition of tables having complicated layouts.

\subsection{DeepFigures}
\label{sec:deepfigure}
Based on our knowledge, DeepFigures \cite{kim2008extracting} is the biggest dataset publicly available to perform the task of table detection. The dataset contains over 1.4 million documents along with their corresponding bounding boxes of tables and figures. The authors leverage the scientific articles available online on the arXiv and PubMed databases to develop the dataset. The ground truth of the dataset \footnote{https://s3-us-west-2.amazonaws.com/ai2-s2-research-public/deepfigures/jcdl-deepfigures-labels.tar.gz} is available in XML format. As highlighted in Table \ref{tab:dataset}, this dataset only contains bounding boxes for the tables. In order to completely exploit deep neural networks for the problem of table detection, this large-scale dataset can be treated as a base dataset to implement closer domain fine-tuning techniques.

\subsection{RVL-CDIP (Subset)}
\label{sec:rvlcdip}
The RVL-CDIP (Ryerson Vision Lab Complex Document Information Processing) \cite{harley2015evaluation} is a renowned dataset in the document analysis community. It contains 400 thousand images equally distributed into 16 classes. \textit{Pau et al.} \cite{riba2019table} leverages the RVL-CDIP dataset by annotating its 518 invoices. The dataset \footnote{https://zenodo.org/record/3257319} has been made publicly available for the task of table detection. The dataset has only annotations for the tabular boundaries as mentioned in Table \ref{tab:dataset}. This subset of the actual RVL-CDIP dataset \cite{harley2015evaluation} is an important contribution for evaluating table detection systems specifically designed for invoice document images. 

\subsection{PubTabNet}
\label{sec:pubtabnet}
PubTabNet is another dataset published in December 2019 by Zhong \textit{et al.} \cite{zhong2019image}. PubTabNet\footnote{https://github.com/ibm-aur-nlp/PubTabNet} is currently the largest publicly available dataset that contains over 568 thousand images with their corresponding structural information of tables and content present in each cell. This dataset is created by collecting scientific articles from PubMed Central$^{TM}$ Open Access Subset (PMCOA). The ground truth format of this dataset is in HTML which can be useful for web applications. The authors are confident that this dataset will boost the performance of information extraction systems in the table and they are also planning to publish ground truth for the respective table cells in the future. The important information for the dataset is summarized in Table \ref{tab:dataset}. Along with the TABLE2LATEX-450K dataset \cite{deng2019challenges}, PubTabNet \cite{zhong2019image} allows researchers the independence of training complete parameters of the deep neural networks on the task of table structure extraction or table recognition.

\subsection{IIIT-AR-13K}
\label{sec:iitar}
Recently, Mondal et al. \cite{mondal2020iiit} contributed to the community of graphical page object detection by introducing a novel dataset known as \textit{IIT-AR-13K}. The authors generated this dataset by collecting publicly available annual reports written in English and other languages. The authors claim that this is the largest manually annotated dataset published for solving the problem of graphical page object detection. Apart from the tables, the dataset includes annotations for figures, natural images, logos, and signatures. The publishers of this dataset have provided the train, validation, and test splits for various tasks of page object detection. For table detection, 11000 samples are used for training, whereas 2000 and 3000 samples are assigned for validation and testing purposes, respectively.

\begin{figure}[ht]
    \includegraphics[width = \columnwidth]{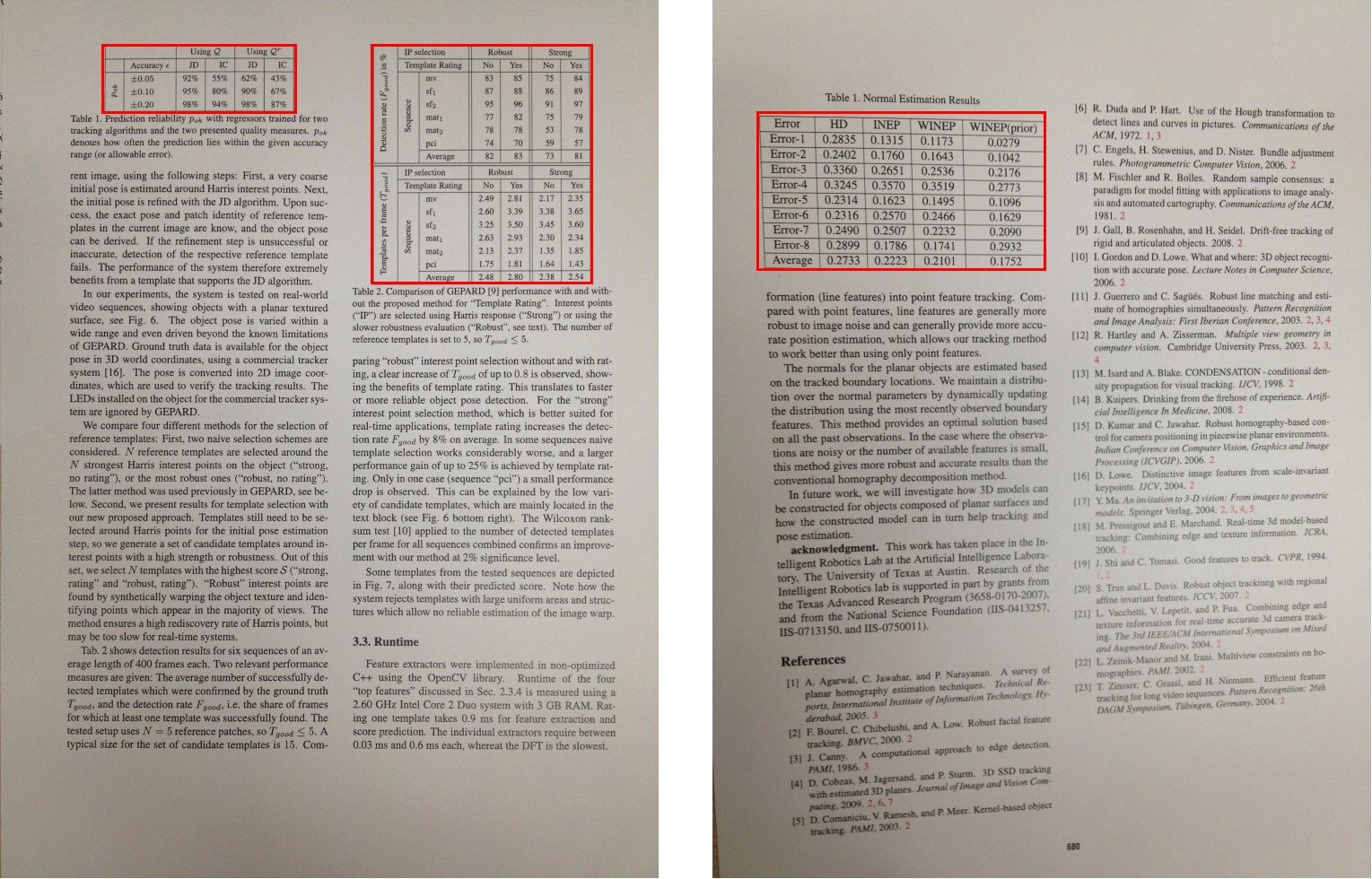}
    \caption{Examples of real camera-captured images taken from the CamCap dataset \cite{seo2015junction} which is explained in Section \ref{sec:CamCap}. The red boundaries represent the tabular region.}
    \label{fig:cam_cap}
\end{figure}

\subsection{CamCap}
\label{sec:CamCap}
CamCap is the last dataset which we have included in this survey consists of the camera-captured images. This dataset is proposed by Seo \textit{et al.}\cite{seo2015junction}. It contains only 85 images (38 tables on curved surfaces having 1295 cells and 47 tables on the planar surfaces consisting of 1162 cells). Figure \ref{fig:cam_cap} contains few samples from this dataset illustrating the challenges. The proposed dataset is publicly available and can be utilized for the task of table detection and table structure recognition as summarized in Table \ref{tab:dataset}. In order to assess the robustness of table detection methods on camera-captured document images, this dataset is an important contribution. 

\begin{figure}[ht]
    \includegraphics[width = \columnwidth]{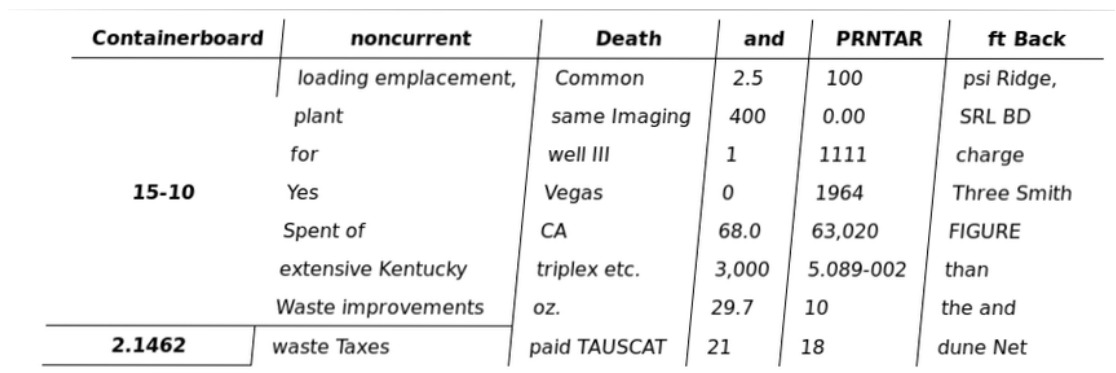}
    \caption{Example of a synthetically created camera captured image by linear perspective transform method \cite{qasim2019rethinking}.}
    \label{fig:syn_cam}
\end{figure}

It is important to mention that Qasim \textit{et al.} \cite{qasim2019rethinking} published a method to synthetically create camera captured images from the UNLV dataset. An instance of a synthetically created camera-captured image is depicted in Figure \ref{fig:syn_cam}.

\section{Evaluation}
\label{sec:evaluation} 
In this section, we will cover the well known evaluation metrics along with the exhaustive evaluation comparisons of all the quoted methodologies from Section \ref{sec:methodology}.

\subsection{Evaluation Metrics}
Before throwing some light on the performance evaluation, it is appropriate to talk about the evaluation metrics first which are adopted to assess the performances of discussed approaches.

\subsubsection{Precision}
\begin{figure}[ht]
    \includegraphics[width = \columnwidth]{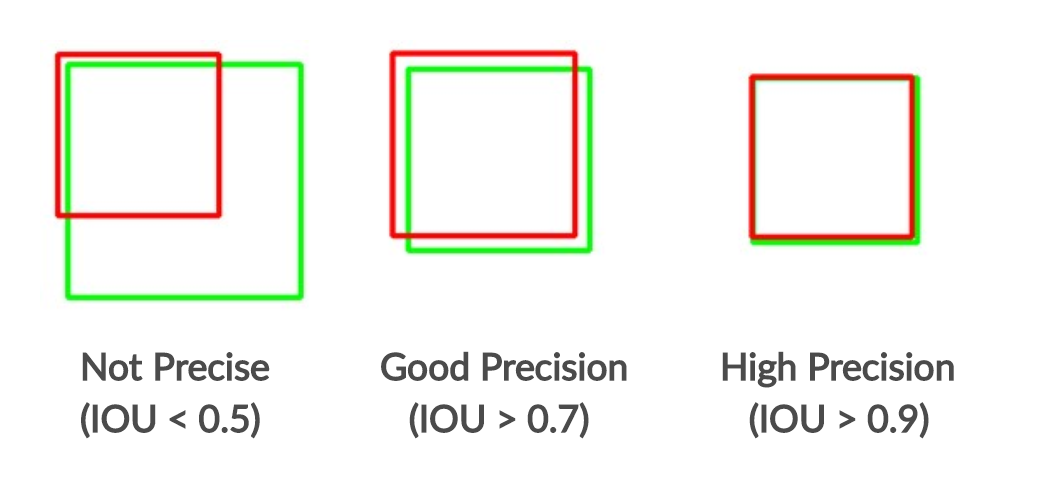}
    \caption{Example of precision in object detection problems where the IOU threshold is set to 0.5. The leftmost case will not be counted as precise whereas the other two predictions are precise because their IOU value is greater than 0.5. Green color represents the ground truth and red color depicts the predicted bounding boxes.}
    \label{fig:precision}
\end{figure}

Precision \cite{powers2020evaluation} is defined as the percentage of a predicted region that belongs to the ground truth. An illustration of different types of precision is explained in Figure \ref{fig:precision}. The formula for precision is mentioned below :
\begin{equation}
\frac{\text{Predicted area in ground truth}} {\text{Total area of predicted region}}
 = \frac{\text{TP}}{\text{TP $+$ FP}}
\end{equation}

\begin{figure}[ht]
    \includegraphics[width = \columnwidth]{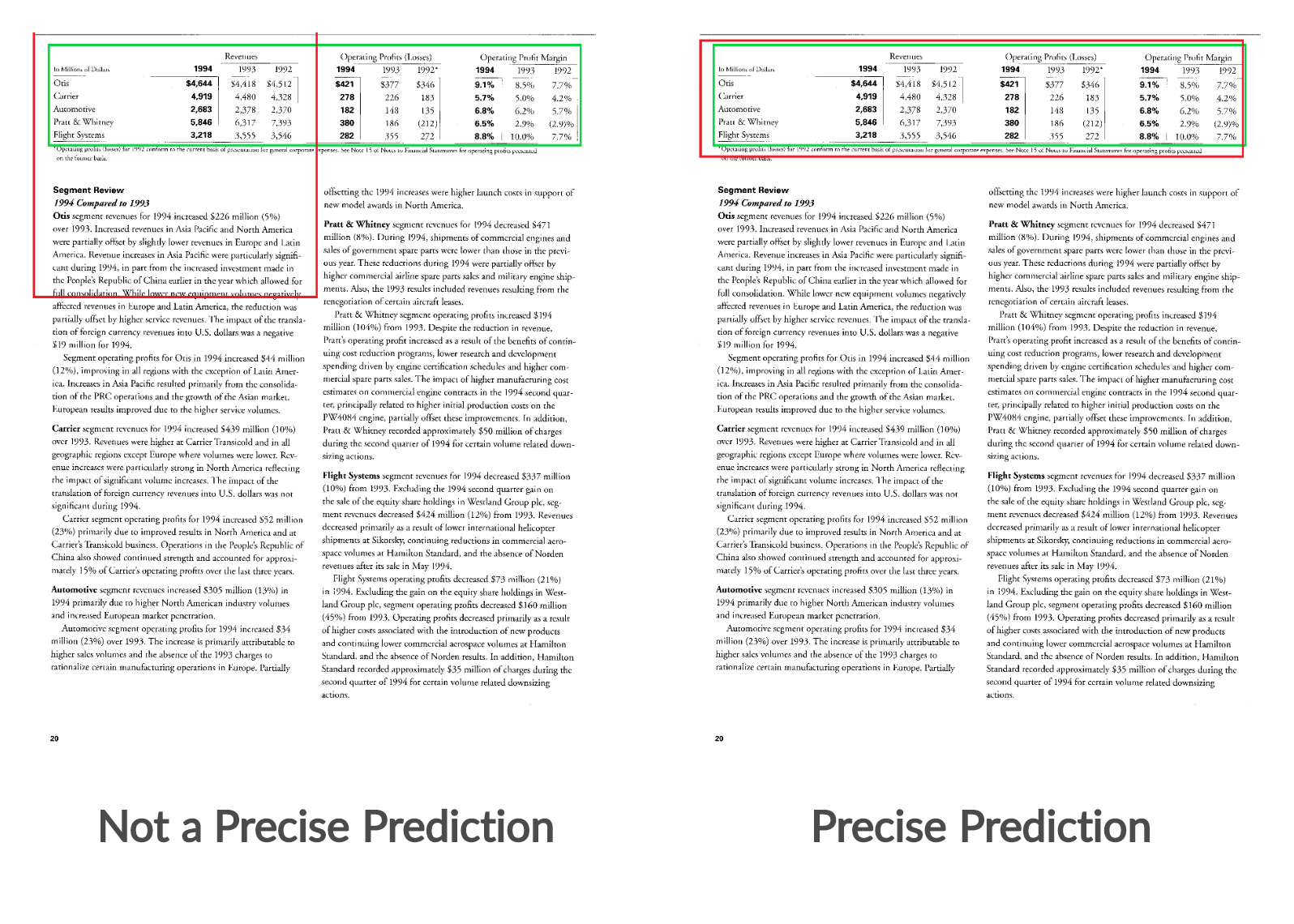}
    \caption{Example of precision in reference to the task of table detection. The green color represents the ground truth whereas the red color depicts the predicted tabular area. In the first case, the prediction is not a precise one because IOU between the predicted bounding box and the ground truth is less than 0.5. The table prediction on the right side is precise because it covers an almost complete tabular area.}
    \label{fig:tbl_det_precision}
\end{figure}

\subsubsection{Recall}
Recall \cite{powers2020evaluation} is calculated as the percentage of ground truth region that is present in the predicted region. The formula for recall is explained as follows : 
\begin{equation}
\frac{\text{Ground truth area in predicted region}} {\text{Total area of ground truth region}}
 = \frac{\text{TP}}{\text{TP $+$ FN}}
 \end{equation}
 
\subsubsection{F-Measure}
F-measure \cite{powers2020evaluation} is calculated by taking the harmonic mean of precision and Recall . The formula for F-measure is :
\begin{equation}
\frac{\text{$2 \times$ Precision $\times$ Recall}} {\text{Precision $+$ Recall}}
 \end{equation}
 
\subsubsection{Intersection Over Union (IOU)}
Intersection over union \cite{blaschko2008learning} is an important evaluation metric which is regularly employed to determine the performance of object detection algorithms. It is the measure of how much the predicted region is overlapping with the actual ground truth region. It is defined as follows :
\begin{equation}
\frac{\text{Area of Overlap region}} {\text{Area of Union region}}
 \end{equation}

\subsubsection{BLEU score}
BLEU (Bilingual Evaluation Understudy) \cite{papineni2002bleu} is an evaluation method utilized to compare in various machine translation problems. After comparing the predicted text with the actual ground truth, a score is calculated. The BLEU metric scores the prediction from 0 to 1 where 1 is the optimal score for the predicted text.

\subsection{Evaluations for Table Detection}

The problem of table detection is to distinguish the tabular area in the document image and regress the coordinates of a bounding box that is classified as a tabular region. Table \ref{tab:detection} explains the performance comparison of various table detection methods that have been discussed in detail in Section \ref{sec:table_detect}. In most of the cases, the performance of the table detection methods is evaluated on ICDAR-2013 \cite{icdar13}, ICDAR-2017-POD \cite{icdar17} and UNLV \cite{shahab2010open} datasets.

The threshold of Intersection Over Union (IOU) for calculating precision and recall is also defined in Table \ref{tab:detection}. Figure \ref{fig:tbl_det_precision} explains the definition of a precise and imprecise prediction in reference to the task of table detections. Results having the highest accuracies in all respective datasets are highlighted. It is crucial to mention that some of the approaches have not quoted the threshold value for IOU; however, they have compared their results with other methods where the threshold value is defined. Hence, we have considered the same threshold value for those procedures.

We could not incorporate the results of the literature presented by \textit{Martin et al.} \cite{holevcek2019table} because they have not adopted any standard dataset for the comparison, and compared their novel method with logistic regression \cite{kleinbaum2002logistic}. The results demonstrate that their model has surpassed the logistic regression method.
%HUGE TABLE HAVING ALL RESULTS FROM TABLE DETECTION%
\newcolumntype{d}{X}
\newcolumntype{t}{>{\hsize=.75\hsize}X}
\newcolumntype{u}{>{\hsize=.45\hsize}X}
\newcolumntype{v}{>{\hsize=.2\hsize}X}
\newcolumntype{k}{>{\hsize=.5\hsize}X}
\begin{table*}
\centering
\caption{Table Detection Performance Comparison. The double horizontal line partitions the results obtained on various datasets. Outstanding results in all the respective datasets are highlighted. For the ICDAR-2019 dataset \cite{icdar19}, all of the three approaches are not directly comparable to each other because they report F-Measure on different IOU thresholds. Hence, results on ICDAR-2019 dataset are not highlighted. }
\normalsize
\setlength\extrarowheight{5pt}
% default value: 6pt{p{0.1\textwidth}p{0.8\textwidth}}
% \begin{tabularx}{\textwidth}{|C|C|C|C|C|C|C|} 
\begin{tabularx}{\textwidth}{t|v|c|c|c|c|c|d}
\specialrule{.2em}{.1em}{.1em} 
\captionsetup{font=scriptsize}
\textbf{Literature}&
\textbf{Year}&
\textbf{Dataset}&
\textbf{IOU }&
\textbf{Precision}&
\textbf{Recall}&
\textbf{F-Measure}&
\textbf{Method}\\
% \hline
\specialrule{.2em}{.1em}{.1em}
\small
    \textit{ Gelani et al.} \cite{gilani2017table}& \small 
    2017& \footnotesize 
    UNLV& \small 
    0.9& \small 

    \textbf{82.3}& \small 
    \textbf{90.6}& \small 
    \textbf{86.3}& \small 
    \footnotesize {Faster R-CNN (Section \ref{sec:fasterrcnn})} \\
    \hline
    \small \textit{García et al.} \cite{casado2020benefits}& \small 
    2019& \footnotesize 
    UNLV& \small 
    0.9& \small 
    48.0& \small 
    49.0& \small 
    49.0& \small 
    \footnotesize 
    YOLO (Section \ref{sec:YOLO})\\
    \hline
    \small \textit{DeCNT} \cite{siddiqui2018decnt}& \small 
    2018& \footnotesize 
    UNLV& \small 
    0.5& \small 
    78.6& \small 
    74.9& \small 
    76.7& \small 
    \footnotesize 
    Deformable Convolutions (Section \ref{sec:deformconv})\\
    \specialrule{.2em}{.1em}{.1em} 
    \specialrule{.2em}{.1em}{.1em}
    \small \textit{DeepDeSRT} \cite{schreiber2017deepdesrt}& \small 
    2017& \footnotesize 
    ICDAR-2013& \small 
    0.5& \small 
    97.4& \small 
    96.1& \small 
    96.7& \footnotesize 
    Faster R-CNN (Section \ref{sec:fasterrcnn})\\
    \hline
    \small \textit{Kavasidis et al.} \cite{kavasidis2018saliency}& \small 
    2018& \footnotesize 
    ICDAR-2013& \small 
    0.5& \small 
    97.5& \small 
    98.1& \small 
    97.8& \footnotesize 
    Semantic Image Segmentation ( Section \ref{sec:sis} )\\
    \hline
    \small \textit{DeCNT} \cite{siddiqui2018decnt}& \small 
    2018& \footnotesize 
    ICDAR-2013& \small 
    0.5& \small 
    99.6& \small 
    {99.6}& \small 
    {99.6}& \footnotesize 
    Deformable Convolutions (Section \ref{sec:deformconv})\\
    \hline
    \small \textit{Huang et al.} \cite{huang2019yolo}& \small 
    2015& \footnotesize 
    ICDAR-2013& \small 
    0.5& \small 
    {100}& \small 
    94.9& \small 
    97.3& \footnotesize 
    YOLO (Section \ref{sec:YOLO})\\
    \hline
    \small \textit{TableBank} \cite{li2020tablebank}& \small  
    2019& \footnotesize 
    ICDAR-2013& \small 
    0.5& \small 
    96.2& \small 
    96.2& \small 
    96.2 & \footnotesize 
    Faster R-CNN (Section \ref{sec:fasterrcnn})\\
    \hline
    \small \textit{TableNet} \cite{paliwal2019tablenet}& \small  
    2019& \footnotesize 
    ICDAR-2013& \small 
    0.5& \small 
    96.3& \small 
    96.9& \small 
    96.6& \footnotesize 
    Fully Convolutional Networks (Section \ref{sec:fcn})\\
    \hline
    \small \textit{CascadeTabNet} \cite{prasad2020cascadetabnet}& \small  
    2020& \footnotesize 
    ICDAR-2013& \small 
    0.5& \small 
    \textbf{100}& \small 
    \textbf{100}& \small 
    \textbf{100}& \footnotesize 
    Cascade Mask R-CNN \newline (Section \ref{sec:cascade})\\
    \hline
    \small \textit{GTE} \cite{zheng2021global}& \small  
    2021& \footnotesize 
    ICDAR-2013& \small 
    0.5& \small 
    -& \small 
    -& \small 
    95.7& \footnotesize 
    Object Detection (Section \ref{sec:cascade})\\
    \hline
    \small \textit{CDeC-Net} \cite{agarwal2020cdec}& \small  
    2020& \footnotesize 
    ICDAR-2013& \small 
    0.5& \small 
    94.2& \small 
    99.3& \small 
    96.8& \footnotesize 
    Cascade Mask R-CNN \newline (Section \ref{sec:cascade})\\
    \hline
    \small \textit{García et al.} \cite{casado2020benefits}& \small 
    2019& \footnotesize 
    ICDAR-2013& \small 
    0.6& \small 
    70.0& \small 
    97.0& \small 
    81.0& \small 
    \footnotesize 
    Mask R-CNN (Section \ref{sec:mask-yolo-ssd})\\
    \specialrule{.2em}{.1em}{.1em} 
    \specialrule{.2em}{.1em}{.1em}
    \small \textit{Li et al.} \cite{li2019gan}& \small 
    2019& \footnotesize 
    ICDAR-2017& \small 
    0.6& \small 
    94.4& \small 
    94.4& \small 
    94.4& \footnotesize 
    GANs (Section \ref{sec:gan})\\
    \hline
    \small \textit{DeCNT} \cite{siddiqui2018decnt}& \small 
    2018& \footnotesize 
    ICDAR-2017& \small 
    0.6& \small 
    {97.1}& \small 
    {96.5}& \small 
    {96.8}& \footnotesize 
    Deformable Convolutions (Section \ref{sec:deformconv})\\
    \hline
    \small \textit{Huang et al.} \cite{huang2019yolo}& \small 
    2015& \footnotesize 
    ICDAR-2017& \small 
    0.6& \small 
    \textbf{97.8}& \small 
    \textbf{97.2}& \small 
    \textbf{97.5}& \footnotesize 
    YOLO (Section \ref{sec:YOLO})\\
    \hline
    \small \textit{Sun et al.} \cite{sun2019faster}& \small  
    2019& \footnotesize 
    ICDAR-2017& \small 
    0.6& \small 
    94.3& \small 
    95.6& \small 
    94.5& \footnotesize 
    Faster R-CNN (Section \ref{sec:fasterrcnn})\\
    \hline
    \small \textit{García et al.} \cite{casado2020benefits}& \small 
    2019& \footnotesize 
    ICDAR-2017& \small 
    0.6& \small 
    92.0& \small 
    87.0& \small 
    89.0& \small 
    \footnotesize 
    Retina Net (Section \ref{sec:mask-yolo-ssd})\\
    \hline
    \small \textit{CDeC-Net} \cite{agarwal2020cdec}& \small  
    2020& \footnotesize 
    ICDAR-2017& \small 
    0.6& \small 
    89.9& \small 
    96.9& \small 
    93.4& \footnotesize 
    Cascade Mask R-CNN \newline (Section \ref{sec:cascade})\\
    \specialrule{.2em}{.1em}{.1em}
    \specialrule{.2em}{.1em}{.1em}
    
    %% Table DETECTION ICDAR 19 BEGIN 
    \small \textit{CascadeTabNet} \cite{prasad2020cascadetabnet}& \small  
    2020& \footnotesize 
    ICDAR-2019& \small 
    0.6& \small 
    -& \small 
    -& \small 
    94.3& \footnotesize 
    Cascade Mask R-CNN \newline (Section \ref{sec:cascade})\\
    \hline
    \small \textit{CDeC-Net} \cite{agarwal2020cdec}& \small  
    2020& \footnotesize 
    ICDAR-2019& \small 
    0.6& \small 
    93.9& \small 
    98.0& \small 
    95.9& \footnotesize 
    Cascade Mask R-CNN \newline (Section \ref{sec:cascade})\\
    \hline
    \small \textit{GTE} \cite{zheng2021global}& \small  
    2021& \footnotesize 
    ICDAR-2019& \small 
    0.8& \small 
    96.0& \small 
    95.0& \small 
    95.5& \footnotesize 
    Object Detection (Section \ref{sec:cascade})\\
    \specialrule{.2em}{.1em}{.1em}
    \specialrule{.2em}{.1em}{.1em}
    %% Table DETECTION ICDAR 19 END
    \small \textit{Riba et al.} \cite{riba2019table}& \small  
    2019& \footnotesize 
    RVL-CDIP& \small 
    0.5& \small 
    15.2& \small 
    36.5& \small 
    21.5& \small  \footnotesize 
    Graph Neural Network (Section \ref{sec:gnn})\\
    \specialrule{.2em}{.1em}{.1em}

\end{tabularx} 
\label{tab:detection}
\end{table*}

% tABLE DETECTION RESULT TABLE ENDS HERE **********%

\begin{figure}[ht]
     \includegraphics[width = \columnwidth]{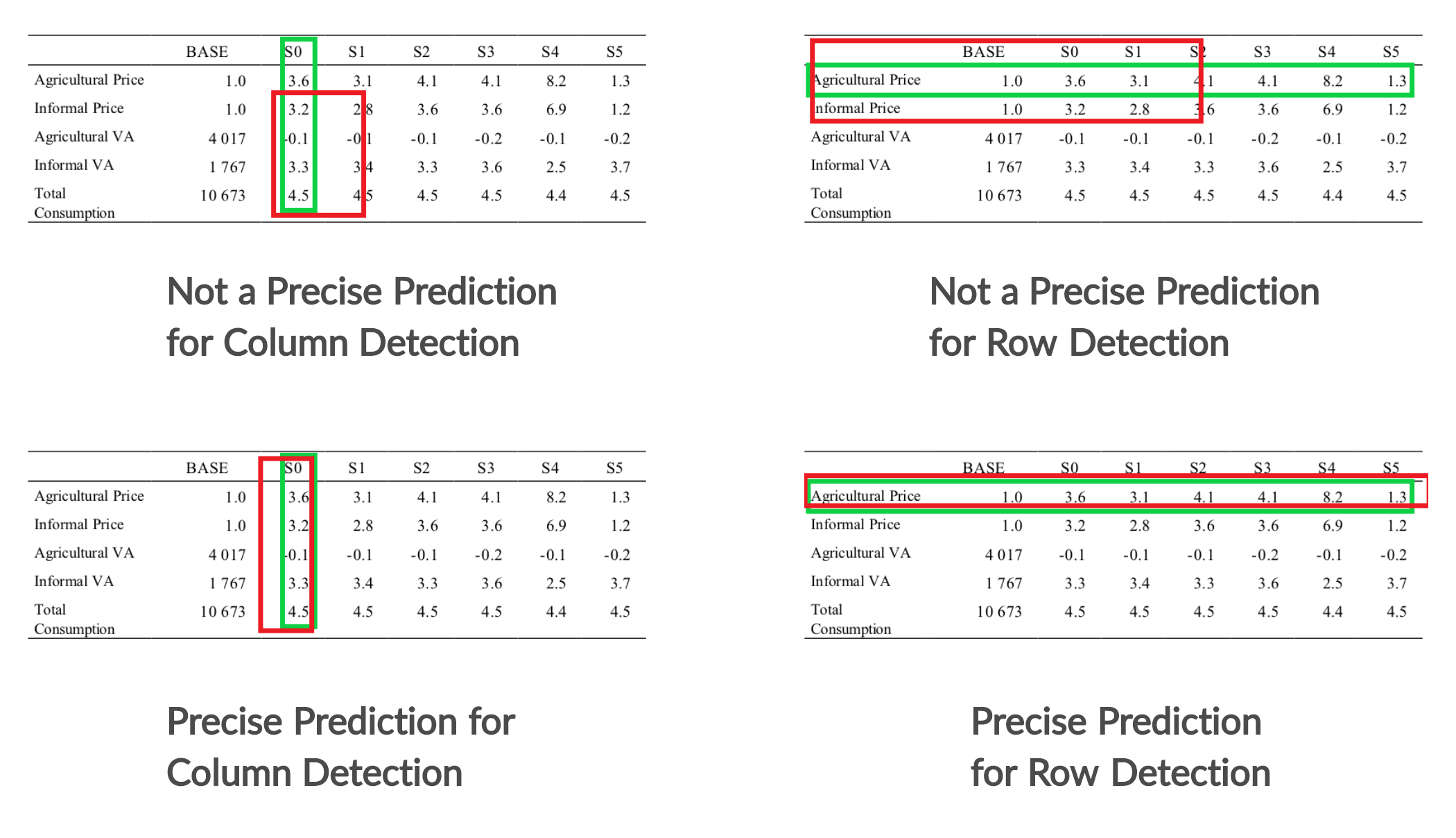}
    \caption{Example of precision in reference to the task of table structural segmentation. Green color represents the ground truth whereas the red color depicts the predicted bounding boxes. For simplicity, precision for detection of rows and columns are shown separately. The IOU threshold in the shown examples is considered as 0.5.}
    \label{fig:tbl_str_prec}
\end{figure}

%HUGE TABLE HAVING ALL RESULTS FROM TABLE STRUCTURAL SEGMENTATION%

\begin{table*}
\centering

\caption{Table Structural Segmentation Performance. Outstanding results are highlighted. Results in the last two rows are not directly comparable with other methods because PDF files are employed instead of document images. }
\normalsize
\setlength\extrarowheight{5pt}
% default value: 6pt{p{0.1\textwidth}p{0.8\textwidth}}
% \begin{tabularx}{\textwidth}{|C|C|C|C|C|C|C|} 
\begin{tabularx}{\textwidth}{l|v|u|c|c|c|c|t}
\specialrule{.2em}{.1em}{.1em} 
\captionsetup{font=scriptsize}
\multirow{2}{*}{\textbf{Literature}}&
\multirow{2}{*}{\textbf{Year}}&
\multirow{2}{*}{\textbf{Dataset}}&
\multirow{2}{*}{\textbf{IOU }}&
\multicolumn{3}{c|}{\textbf{Average Row/Column}}&
\multirow{2}{*}{\textbf{Method}}\\
\cline{5-7}
& & & & \textbf{Precision}&
\textbf{Recall}&
\textbf{F-Measure}\\
\specialrule{.2em}{.1em}{.1em} 
% \hline
\small \textit{DeepDeSRT} \cite{schreiber2017deepdesrt}& \small 
2017& \small 
ICDAR-2013& \small 
0.5& \small 
95.9& \small 
87.3& \small 
91.4& \footnotesize 
Fully Convolutional Networks (Section \ref{sec:fcn_seg})\\
\hline
\small \textit{DeepTabStR} \cite{siddiqui2019deeptabstr}& \small 
2019& \small 
ICDAR-2013& \small 
0.5& \small 
93.1& \small 
{93.0}& \small 
92.9& \footnotesize 
Deformable Convolutions  (Section \ref{sec:ddc})\\
\hline
\small \textit{ReS${^2}$TIM} \cite{xue2019res2tim}& \small  % Cell level
2019& \small 
ICDAR-2013& \small 
0.5& \small 
92.6& \small 
44.7& \small 
60.3& \footnotesize 
Distance based weight technique (Section \ref{sec:dbw}) \\
\hline
\small \textit{Siddiqui et al.} \cite{schreiber2017deepdesrt}& \small  
2019& \small 
ICDAR-2013& \small 
0.5& \small 
92.5& \small 
92.7& \small 
92.3 & \footnotesize 
Fully Convolutional Networks (Section \ref{sec:fcn_seg})\\
\hline
\small \textit{Khan et al.} \cite{khan2019table}& \small  
2019& \small 
ICDAR-2013& \small 
0.5& \small 
\textbf{96.9}& \small 
90.1& \small  
{93.3} & \footnotesize 
Bi-directional Recurrent Neural Networks (Section \ref{sec:rnn}) \\
\hline
\small \textit{TableNet} \cite{paliwal2019tablenet}& \small  
2019& \small 
ICDAR-2013& \small 
0.5& \small 
{92.1}& \small 
89.9& \small 
90.1 & \footnotesize 
Fully Convolutional Networks (Section \ref{sec:fcn_seg})\\

%%%%%%%%%%% NEW ENTRIED ADDED HERE %%%%%%%%%%
\hline
\small \textit{Hashmi et al.} \cite{hashmi2021guided}& \small  
2021& \small 
ICDAR-2013& \small 
0.5& \small 
95.4& \small 
\textbf{95.6}& \small 
\textbf{95.5} & \footnotesize 
Object Detection (Section \ref{sec:odstr})\\
\hline
\small \textit{Raja et al.} \cite{khan2019table}& \small  
2020& \small 
ICDAR-2013& \small 
0.5& \small 
92.7& \small 
91.1& \small 
91.9 & \footnotesize  
Object Detection (Section \ref{sec:odstr})\\

%%%%%%%%%%% NEW ENTRIED FINSIHED HERE %%%%%%%%%%

\specialrule{.2em}{.1em}{.1em} 
\specialrule{.2em}{.1em}{.1em} 
\small\textit{Tensmeyer et al. } \cite{tensmeyer2019deep}& \small 
2019& \small 
ICDAR-2013& \small 
0.5& \small 
 95.8& \small 
{94.6}& \small 
{95.2}& \footnotesize 
Dilated Convolutions  (Section \ref{sec:ddc})\\
\hline
\small \textit{GraphTSR} \cite{chi2019complicated}& \small  
2019& \small 
ICDAR-2013& \small 
0.5& \small 
{88.5}& \small 
86.0& \small 
87.2 & \footnotesize 
Fully Convolutional Networks (Section \ref{sec:fcn_seg})\\
\specialrule{.2em}{.1em}{.1em} 
\end{tabularx}

\label{tab:segmentation}
\end{table*} 
%%% Table results with ICDAR-19

\begin{table*}
\centering

\caption{Table Structural Segmentation Performance on the dataset of ICDAR-2019 \cite{icdar19}. For brevity and clarity, these results are separately presented in this table.}
\normalsize
\setlength\extrarowheight{5pt} 
\begin{tabularx}{\textwidth}{c|c|c|c|c|l}
\specialrule{.2em}{.1em}{.1em} 
\captionsetup{font=scriptsize}
\textbf{Literature}&
\textbf{Year}&
\textbf{Dataset}&
% {a \\ bb \\ c}&
\textbf{\shortstack{ \\ F-Measure \\ (0.6) }} &
\textbf{\shortstack{ \\ F-Measure \\ (0.8) }} &
\textbf{Method}\\
\specialrule{.2em}{.1em}{.1em} 
% \hline
\small \textit{CascadeTabNet} \cite{prasad2020cascadetabnet}& \small 
2020& \small 
ICDAR-2019& \small 
\textbf{43.8}& \small 
\textbf{19.0}& 
\footnotesize 
Object detection (Section \ref{sec:odstr})\\
\hline
\small \textit{GTE} \cite{zheng2021global}& \small 
2021& \small 
ICDAR-2019& \small 
38.5& \small 
-& 
\footnotesize Object detection (Section \ref{sec:odstr})\\
\hline
\small \textit{Zou et al.} \cite{zou2020deep}& \small 
2020& \small 
ICDAR-2019& \small 
13.1& 
1.1& 
\footnotesize Fully Convolutional Networks (Section \ref{sec:fcn_seg})\\
\specialrule{.2em}{.1em}{.1em} 
\end{tabularx}

\label{tab:str_recognition_icdar19}
\end{table*} 

% tABLE structure segmentation RESULT TABLE ENDS HERE **********%

Another method by Qasim \textit{et al.} \cite{qasim2019rethinking} which is explained in Section \ref{sec:gnn} did not use any well known dataset to evaluate their approach. However, they have tested their approach on the synthetic dataset by using two types of graph neural networks which are \cite{wang2019dynamic} and \cite{qasim2019learning}. Along with the graph neural networks, a fully convolutional neural network was used to conduct a fair comparison. After an exhaustive evaluation, the fusion of graph neural network and the convolutional neural network has surpassed all the other methods with a perfect matching accuracy of 96.9. The approach which uses only graph neural networks has delivered perfect matching accuracy of 65.6, which still exceeds the accuracy of the method using only fully convolutional neural networks.
% write here about rethinking tabe using GNN %

\subsection{Evaluations for Table Structural Segmentation}
\label{sec:result_segmentation}

The task of table structural segmentation is evaluated based on how accurate the rows or columns of the tables are separated \cite{schreiber2017deepdesrt,siddiqui2019deeptabstr,schreiber2017deepdesrt}. Figure \ref{fig:tbl_str_prec} illustrates the meaning of an imprecise and precise prediction for both of the tasks of the row and column detections. Table \ref{tab:segmentation} summarizes the performance comparison of numerous approaches which have executed the task of table structural segmentation on the ICDAR 2013 table competition dataset \cite{icdar13}. The results with the highest accuracies are highlighted in the table. It is important to mention that apart from the methods mentioned in Table \ref{tab:segmentation}, there are two other approaches which are discussed in section \ref{sec:segmentation}. We could not incorporate their results in Table \ref{tab:segmentation} because the approaches are neither evaluated on any standard dataset nor utilized the standard evaluation metrics. However, their results are explained in the following paragraph.

%HUGE TABLE HAVING ALL RESULTS FROM TABLE RECOGNITION
\begin{table*}
\centering

\caption{Table Recognition Performance. Results mentioned in this table are not directly comparable with each other because different datasets and evaluation metrics have been used.}
\normalsize
\setlength\extrarowheight{5pt} 
\begin{tabularx}{\textwidth}{c|c|c|c|c|l}
\specialrule{.2em}{.1em}{.1em} 
\captionsetup{font=scriptsize}
\textbf{Literature}&
\textbf{Year}&
\textbf{Dataset}&
\textbf{Evaluation Metric}&
\textbf{Score}&
\textbf{Method}\\
\specialrule{.2em}{.1em}{.1em} 
% \hline
\small \textit{Deng et al.} \cite{deng2019challenges}& \small 
2019& \small 
TABLE2LATEX-450K& \small 
BLEU& \small 
40.3& 
\footnotesize 
Encoder Decoder Network (Section \ref{sec:enc-dec})\\
\hline
\small \textit{Zhong et al.} \cite{zhong2019image}& \small 
2020& \small 
PubTabNet& \small  
TEDS \cite{zhong2019image}& \small 
88.3& 
\footnotesize Encoder Dual Decoder Model (Section \ref{sec:enc-dual-dec})\\
\specialrule{.2em}{.1em}{.1em} 
\end{tabularx}

\label{tab:recognition}
\end{table*}

% TABLE RECOGNITION RESULT TABLE ENDS HERE **********%

The creators of the \textit{TableBank} \cite{li2020tablebank} have proposed baseline model for table structure segmentation along with table detection. To examine the performance of their baseline model for table structure recognition on \textit{TableBank} dataset, they have employed the 4-gram BLEU score \cite{papineni2002bleu} as the evaluation metric. The result shows that when their Image-to-Text model is trained on the Word+Latex dataset, it gives the BLEU score of 0.7382 and also generalizes better in all the cases.

\subsection{Evaluations for Table Recognition}
Table recognition consists of both segmenting the structure of tables and extracting the information from the cells. In this section, we will present the evaluations of the couple of approaches that are discussed above in Section \ref{sec:table_rec}.

In the study of challenges in end-to-end neural scientific table recognition, the author {Deng et al.} \cite{deng2019challenges} have tested their image-to-text model on the TABLE2LATEX-450K dataset. The model obtained $32.40\%$ exact match accuracy with a BLEU score of 40.33. The authors have also examined the model that how well it identifies the structure of the table. It has been concluded that the model encounters problems in case of complex structures having multi-column (rows).

Another research by \textit{Zhong et al.} \cite{zhong2019image} has also carried out experiments on the task of table recognition. To evaluate the observations, they have come up with their own evaluation metric called TEDS in which the similarity is calculated using the same tree edit distance proposed by \textit{Pawlik et al.} \cite{pawlik2016tree}. Their Encoder-Dual-Decoder (EDD) model has beaten all the other baseline models with the TEDS score of $88.3\%$ on the PubTabNet dataset. 

The results of both of the discussed methods are summarized in Table \ref{tab:recognition}. It is important to mention that the presented approaches are not directly comparable to each other because of the disparate datasets and evaluation metrics utilized in these techniques.

\section{Conclusion}
\label{sec:conclusion}
Table analysis is a crucial and well-studied problem in the area of document analysis community. The exploitation of deep learning concepts have remarkably revolutionized the problem of table understanding and has set new standards. In this review paper, we have discussed some recent contemporary procedures that have applied the notions of deep learning to progress the task of information extraction from tables in document images. In Section \ref{sec:methodology}, we have explained the approaches that have exploited deep learning to perform table detection, structure segmentation, and recognition. Figure \ref{graph:detect} and Figure \ref{graph:segmentation} illustrate the most and least famous adopted methods for table detection and structure segmentation respectively. We have summarized all the publicly available datasets along with their access information in Table \ref{tab:dataset}.

In Tables \ref{tab:detection}, \ref{tab:segmentation} and \ref{tab:recognition}, we have provided an exhaustive performance comparison of the discussed approaches on various datasets. We have discussed that state-of-the-art methods for table detection on well known publicly available datasets have achieved near perfection results. Once the tabular area is detected, there comes a task of structural segmentation of tables and table recognition subsequently. After examining several recent approaches, we believe that there is still room left for improvement in both of these areas.

\section{Future Work}
\label{sec:futework}

While analyzing and comparing miscellaneous methodologies, we have noticed some aspects that should be highlighted so they can be taken care of in future works. For table detection, one of the most exploited evaluation metrics is IOU \cite{schreiber2017deepdesrt,siddiqui2018decnt}. The majority of approaches that are discussed in this paper have compared their methods with the previous state-of-the-art methods on the basis of precision, recall, and F-measure \cite{powers2020evaluation}. These three metrics are calculated on a specific IOU threshold established by the authors. We strongly believe that the threshold value for IOU needs to be standardized in order to have an impartial comparison. Another important factor that we have seen missing in several research papers is about mentioning the performance time while comparing different methods. In a few cases, semantic segmentation is proven to outperform other methods for table structure segmentation in terms of accuracy. However, the description about execution time is not evident. 

So far, traditional approaches have been exploited to detect tables from the camera-captured document images \cite{seo2015junction}. The power of deep learning methods could be leveraged to improve the state-of-the-art table analysis systems in this domain. Deep learning leverages huge datasets \cite{schreiber2017deepdesrt}. Recently, large publicly available datasets \cite{li2020tablebank,chi2019complicated,zhong2019image} have been published that provide annotations not only for the table structure extraction but also for table detection. We expect that these contemporary datasets will be tested. The results of table segmentation and recognition methods can be further enhanced by exploiting the blend of various deep learning concepts with recently published datasets. To the best of our knowledge, reinforcement learning \cite{mnih2015human, barto1995reinforcement} has not been investigated in the domain of table analysis but some work exists for information extraction from document images \cite{park2020multi}. Nonetheless, it is an exciting and promising future direction for table detection and recognition as well. 

\bibliographystyle{IEEEtran}
\bibliography{access.bib}

\EOD

\end{document}